%% file: main.tex
\documentclass{article}

\usepackage{arxiv}
\usepackage[utf8]{inputenc} % allow utf-8 input
\usepackage[T1]{fontenc}    % use 8-bit T1 fonts
\usepackage{url}            % simple URL typesetting
\usepackage{booktabs}       % professional-quality tables
\usepackage{amsfonts}       % blackboard math symbols
\usepackage{nicefrac}       % compact symbols for 1/2, etc.
\usepackage{microtype}      % microtypography
\usepackage{hyperref}
\usepackage{lipsum}         % Can be removed after putting your text content
\usepackage{graphicx}
\usepackage{doi}
\usepackage{enumerate}% http://ctan.org/pkg/enumerate
\usepackage{amsmath}
\usepackage{bbm}
\usepackage{booktabs}
\usepackage[table,xcdraw]{xcolor}
\usepackage{multirow}
\usepackage{svg}
\usepackage{cleveref}       % smart cross-referencing
\usepackage{algorithm}
\usepackage{algpseudocode}

\usepackage[style=ieee, isbn=false]{biblatex}
\AtEveryBibitem{
    \clearfield{urlyear}
    \clearfield{urlmonth}
}

\title{Fault Detection in New Wind Turbines with Limited Data by Generative Transfer Learning}
\date{}

\newif\ifuniqueAffiliation

\usepackage{authblk}

\author[1, 2]{%
	Stefan Jonas\thanks{\texttt{stefan.jonas@bfh.ch}}}%

\author[1,3]{%
	Angela Meyer}%

\affil[1]{School of Engineering and Computer Science, Bern University of Applied Sciences, Biel, Switzerland}
\affil[2]{Faculty of Informatics, Università della Svizzera italiana, Lugano, Switzerland}
\affil[3]{Department of Geoscience and Remote Sensing, Delft University of Technology, Delft, The Netherlands}

\renewcommand{\headeright}{}
\renewcommand{\undertitle}{}
\renewcommand{\shorttitle}{}

\begin{document}
\maketitle

\begin{abstract}
Intelligent condition monitoring of wind turbines is essential for reducing downtimes. Machine learning models trained on wind turbine operation data are commonly used to detect anomalies and, eventually, operation faults. However, data-driven normal behavior models (NBMs) require a substantial amount of training data, as NBMs trained with scarce data may result in unreliable fault detection. To overcome this limitation, we present a novel generative deep transfer learning approach to make SCADA samples from one wind turbine lacking training data resemble SCADA data from wind turbines with representative training data. Through CycleGAN-based domain mapping, our method enables the application of an NBM trained on an existing wind turbine to a new one with severely limited data. We demonstrate our approach on field data mapping SCADA samples across 7 substantially different WTs. Our findings show significantly improved fault detection in wind turbines with scarce data. Our method achieves the most similar anomaly scores to an NBM trained with abundant data, outperforming NBMs trained on scarce training data with improvements of +10.3\% in F1-score when 1 month of training data is available and +16.8\% when 2 weeks are available. The domain mapping approach outperforms conventional fine-tuning at all considered degrees of data scarcity, ranging from 1 to 8 weeks of training data. The proposed technique enables earlier and more reliable fault detection in newly installed wind farms, demonstrating a novel and promising research direction to improve anomaly detection when faced with training data scarcity. 
\end{abstract}

\begin{figure}[h!]
    \centering
    \includegraphics[width=0.95\textwidth]{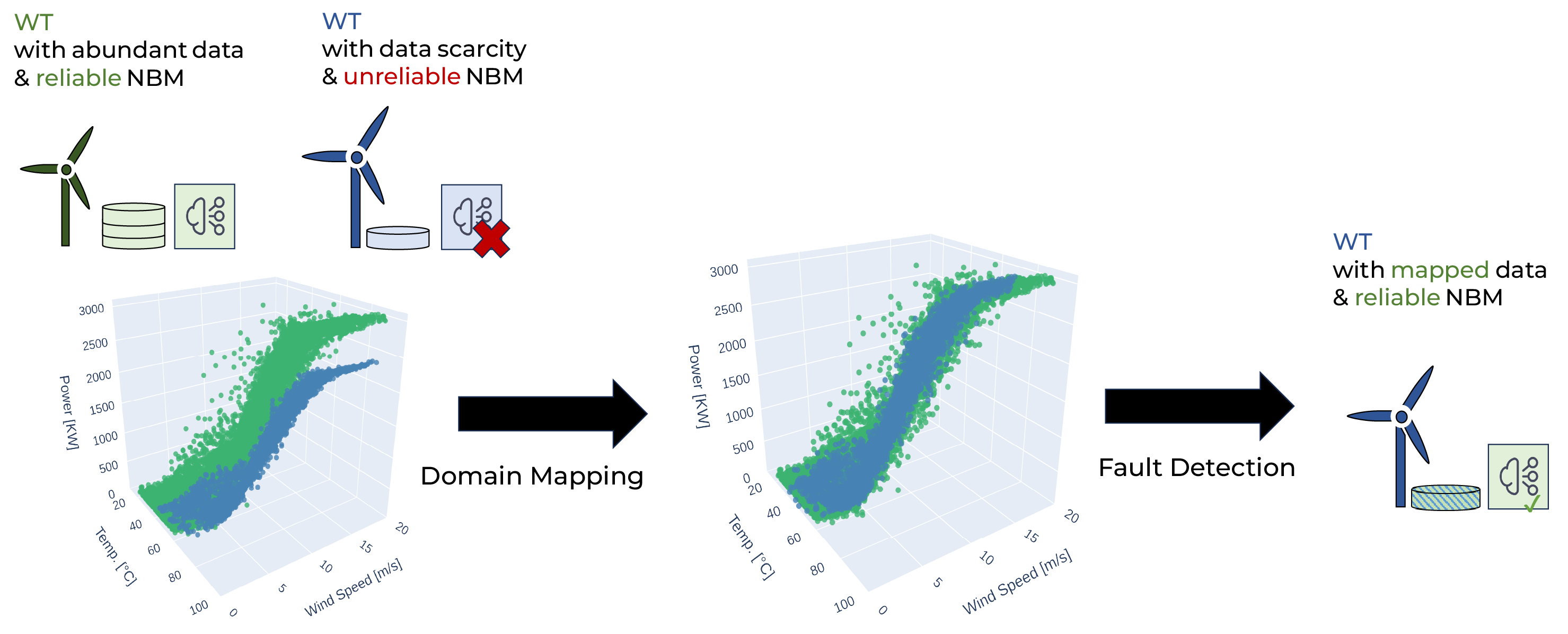}
    %\caption{a nice plot}
    %\label{fig:mesh1}
\end{figure}
\newpage

% keywords can be removed
\keywords{wind turbine \and fault detection \and deep learning \and transfer learning \and generative domain adaptation \and domain mapping \and data scarcity \and anomaly detection}

%%%% --- INTRODUCTION ----- 
%%%% --- INTRODUCTION ----- 
%%%% --- INTRODUCTION ----- 
%%%% --- INTRODUCTION ----- 

\section{Introduction}
Wind energy plays an essential role in the global shift towards renewable energy. With the increasing role of wind energy, improving the reliability of wind turbines (WTs) is of crucial importance \cite{faulstichWindTurbineDowntime2011}. Supervisory control and data acquisition (SCADA) data has facilitated the application of deep learning methods for data-driven condition monitoring tasks in WTs \cite{helbingDeepLearningFault2018a, tautz-weinertUsingSCADAData2017}. Classification models represent a major share of these models \cite{stetcoMachineLearningMethods2019}. However, these supervised classification methods require thoroughly labeled WT maintenance records and a fault database containing SCADA data for a wide variety of fault types, which is often unavailable in practice. Normal behavior models (NBMs) \cite{helbingDeepLearningFault2018a, meyerMultitargetNormalBehaviour2021} can overcome this limitation. NBMs are based on anomaly detection and model the fault-free operation behavior of WT components. As opposed to classification models, NBMs only require a training dataset comprising historical, fault-free operation measurements. Deviations from the expected value obtained from the NBM at test time, i.e., when the trained model is used in an operational context, can be viewed as an anomaly score, with continuous deviations indicating possible incipient faults. This setting of training data from only assumed normal operation falls under unsupervised anomaly detection, a category that is in literature also sometimes referred to as semi-supervised \cite{ruff_review}. 

A major limitation of NBMs is that they require a large training set that is representative of the various WT operation and environmental conditions. Training NBMs with scarce or non-representative data (for instance, only records at low wind speeds) may lead to unreliable fault detection models \cite{grataloupWindTurbineCondition2024, jenkelPrivacyPreservingFleetWideLearning2023}. Large training sets are sometimes unavailable as scarcity of representative training data can occur, for instance, in newly installed WTs, after maintenance, part replacements, or as a result of aging, rendering previously collected data unrepresentative of the future WT operational states. Data scarcity is further exacerbated by the lack of data sharing in the wind industry, which is inhibiting research and development \cite{kusiakRenewablesShareData2016}. A scarcity in training data poses a serious challenge for WT fault detection tasks.

Transfer learning and, in particular, domain adaptation, aims to overcome the limitations caused by data scarcity by transferring knowledge and adapting models from data of a different but related domain \cite{panSurveyTransferLearning2010, wilsonSurveyUnsupervisedDeep2020}. In our fault detection context, domains are represented by a WT and its associated data, knowledge, and NBMs. The goal is to use a source domain where data is abundantly available, represented by a WT with abundant data, to improve a task on a target domain, represented by a WT with scarce training data. Domain adaptation has shown success in numerous WT-related fault classification tasks for classifying unlabeled data (e.g., \cite{renNewWindTurbine2019, xieInvestigationDeepTransfer2023, zhuAnomalyDetectionCondition2022}). However, domain adaptation for unsupervised anomaly detection has hardly been investigated \cite{yangAnomalyDetectionDomain2023}. This is likely a result of multiple challenges faced in applying domain adaptation to unsupervised anomaly detection. First, it usually involves time series data as opposed to the emphasis on image data. Further, by definition, only normal data is available for training anomaly detection models, excluding supervision-based methods. Moreover, unlike the typical setting with abundant but unlabeled target domain data, we are faced with data scarcity in the target domain (i.e., the target wind turbine). Available studies using transfer learning or domain adaptation for improving unsupervised anomaly detection through NBMs are limited to fine-tuning \cite{schroderUsingTransferLearning2022} or simple corrections of the NBM \cite{zgraggenTransferLearningApproaches2021}. 
 
Adapting and transferring knowledge across WTs is essential. Detecting faults in a data-scarce WT by employing an NBM trained on another data-rich WT is generally not feasible. Since the WTs vary in, for instance, power generation and drive train characteristics, the NBM trained on a different WT will predict an incorrect expected state due to the varying characteristics. However, an NBM from another WT exhibiting a minimal domain shift \cite{9238468} might possibly be used to detect faults in the WT with scarce training data. That is, if the 2 WT's test data distributions would match. This would be the case if the two WTs shared the same specifications, operational behavior, and weather conditions. To this end, our work demonstrates how to make the WTs resemble each other by transforming their SCADA data into one another. We employ domain mapping \cite{wilsonSurveyUnsupervisedDeep2020} to map SCADA samples from one WT to resemble the corresponding SCADA data of another WT. Through this mapping, we can enable fault detection using another WT's NBM. A critical component herein is that the mapping should preserve the sample's content when translating it to another domain. For example, a SCADA measurement capturing a WT running idly should remain in an idle state, and critically, anomalous behavior should be mapped to anomalous behavior across WT domains. 

Jin et al.  \cite{jinConditionMonitoringWind2023} proposed a generative neural network that synthesizes SCADA data resembling another WT but did not employ or consider any content preservation. Without explicitly enforcing a preservation of WT operational states, as employed in our study, anomalies may be mapped to healthy states, ultimately leading to unreliable fault detection. Pattnaik et al. \cite{pattnaikCycleGANBasedUnsupervised2022} presented a content-preserving CycleGAN \cite{zhuUnpairedImagetoImageTranslation2017}-based domain mapping approach for mapping industrial time series data across motors for bearing fault classification. While their work demonstrated the promising potential of mapping data between different but related machines, it followed the usual domain adaptation workflow of abundant but unlabeled data for classification, whereas our study investigates domain mapping for an unsupervised anomaly detection task using limited training data.

We propose a novel fault detection approach based on WT domain mapping with CycleGAN that learns to map SCADA data between the source and the target WT domain in a content-preserving manner. At test time, the target data is mapped to the source domain, thereby enabling fault detection using the source WT’s NBM. To our knowledge, this study is the first to demonstrate a CycleGAN-based framework for unsupervised anomaly detection with time series as well as in wind energy. Our contributions are as follows:

\begin{enumerate}[i)]
  \item Our study is the first to investigate and demonstrate domain mapping for unsupervised anomaly detection with time series for WT fault detection. 
  \item We show how our domain mapping technique can be used for data scarce WTs to improve NBM performance in fault detection applications.
  \item We outline future research directions for domain mapping-based anomaly detection in wind energy.
\end{enumerate}

%%%% --- RELATED WORK ----- 
%%%% --- RELATED WORK ----- 
%%%% --- RELATED WORK ----- 
%%%% --- RELATED WORK ----- 

\section{Related Work}
\label{sec:relwork}

\subsection{Domain adaptation}
A promising approach to overcome data scarcity limitations is transfer learning, particularly domain adaptation \cite{panSurveyTransferLearning2010, wilsonSurveyUnsupervisedDeep2020}. The goal is adapt data, models, or knowledge from one domain with abundant data, the source domain, to a domain affected by some data or label scarcity, the target domain. 

Formally, a domain $D=\{\mathcal{X},P\left(X\right)\}$ consists of a feature space $\mathcal{X}$ and a marginal probability distribution $P(X)$, where $X=\{{x}_1,\ldots,x_n\}\in\mathcal{X}$ \cite{panSurveyTransferLearning2010}. We aim to learn a task $T = \{\mathcal{Y}, P(Y|X)\}$, where $\mathcal{Y}$ is the label space and $P(Y|X)$ is the conditional probability. We differentiate between the source domain $\mathcal{D}_\mathcal{S}$ with its dataset $D_s=\left(x_{1s},y_{1s}\right),\ldots,\left(x_{ns},y_{ns}\right)$ and the target domain $\mathcal{D}_T$ with its corresponding dataset $D_T=\left(x_{1t},y_{1t}\right),\ldots,\left(x_{nt},y_{nt}\right)$, where usually $0<n_t\ll n_s$. In our context, the source domain is represented by a source WT with representative SCADA data ($n_s\gg0$) and the target domain by a target WT with scarce data ($n_t\ll n_s$). Transfer learning aspires to use knowledge extracted from $D_S$ and by $T_S$ to improve learning $T_T$ in the target domain $\mathcal{D}_T\neq\mathcal{D}_\mathcal{S}$. Domain adaptation is a particular type of transductive transfer learning, a setting in which the task remains the same but across different domains \cite{panSurveyTransferLearning2010, wilsonSurveyUnsupervisedDeep2020}.

Much attention has been devoted to unsupervised domain adaptation, a setting with a labeled source domain but an unlabeled target domain, since this is common for vision datasets where acquiring labels is a time intensive and expensive task \cite{wilsonSurveyUnsupervisedDeep2020}. Initial approaches were focused on learning domain-invariant representations of source and target features to enable classification of the unlabeled target domain (e.g., \cite{deepCORAL, ganinAdversarial}).

More recently, generative methods were able to improve performance in unsupervised domain adaptation \cite{liuDeepUnsupervisedDomain2022, 9238468}. These methods extend domain adaptation with a generative component. A main approach is domain mapping, which aims to map (translate) samples from one domain to another. Domain mapping can allow the use of a source domain classifier for unlabeled mapped target data. As an example, we consider datasets of digits. Given unlabeled handwritten digits, no classifier can be trained. But with domain mapping, they can be mapped even without defined pairs to resemble another domain, e.g., computer-generated numbers, for which labels and a classifier are available. An essential challenge is to retain important characteristics of the input, such as its class, to ensure a correct classification. That is, a handwritten digit \textit{7} must remain a \textit{7} when mapped. The underlying unpaired image-to-image translation methods (e.g., \cite{liuUnsupervisedImagetoImageTranslation2017, zhuUnpairedImagetoImageTranslation2017}) are fundamental to achieve this. CycleGAN \cite{zhuUnpairedImagetoImageTranslation2017} enables the translation of unpaired samples into another domain with generative adversarial networks through a cycle-consistency loss enforcing the preservation of content. Domain mapping with CycleGAN has been used in various applications, e.g. in medical settings \cite{palladinoUnsupervisedDomainAdaptation2020}, voice conversion \cite{kanekoParallelDataFreeVoiceConversion2017}, machine fault diagnosis \cite{pattnaikCycleGANBasedUnsupervised2022}, or also for synthetic data augmentation in WT icing detection \cite{chatterjeeDomaininvariantIcingDetection2023}. For a more comprehensive review of unsupervised domain adaptation we refer to \cite{liuDeepUnsupervisedDomain2022, 9238468}. 

\subsection{Domain adaptation for wind turbine fault detection}
The potential of domain adaptation has been demonstrated with time series data from sensors across multiple applications, including non-WT industrial fault detection \cite{pattnaikCycleGANBasedUnsupervised2022, shiDeepUnsupervisedDomain2022, yanComprehensiveSurveyDeep2024}. These applications primarily rely on classification through domain alignment with abundant target domain data lacking fault labels. For WTs specifically, discrepancy-based domain adaptation for fault classification is presented in \cite{renNewWindTurbine2019, xieInvestigationDeepTransfer2023, yueSpatioTemporalFeatureAlignment2024, zhuAnomalyDetectionCondition2022}. Results show significant gains in accuracy on the unlabeled target domains, even when target domain data is scarce \cite{xieInvestigationDeepTransfer2023}. An alternative approach to improve fault classification performance, namely by generative data augmentation, is proposed in \cite{liuNovelTransferLearning2022} to generate missing working conditions for vibration data.

\newpage 
Yet, the available literature for our relevant anomaly detection task using NBMs is scarce. In our unsupervised anomaly detection setting, there is typically a target data scarcity, coupled with no available supervision (e.g., no fault labels), thereby rendering many previously presented domain adaptation techniques unsuitable for this task. Few conventional transfer learning approaches have been proposed: In \cite{schroderUsingTransferLearning2022}, fine-tuning is successfully applied to an NBM pretrained on physics-informed simulation data. While the results show significantly improved performance for when only one month of SCADA data is available, the physics-informed simulation restricts the models to active power NBMs. A simple correction based on linear regression of the source domain NBM output is proposed in \cite{zgraggenTransferLearningApproaches2021}. The proposed correction achieves slightly improved results over fine-tuning but is primarily implemented to avoid overfitting an NBM on a specific season, represented by a target training set comprising considerable 3 months of data.   

The first generative domain adaptation work for WT NBMs is presented by Jin et al. \cite{jinConditionMonitoringWind2023}. Specifically, a generative adversarial network (GAN) is proposed to map samples from a data-scarce WT, comprising two weeks of SCADA data, to a data-rich WT. Like in domain mapping approaches, the GAN is conditioned on target data, that is, the input to the generator are SCADA samples from the scarce target domain, rather than random noise. In doing so, the target domain input is mapped to samples matching the marginal probability distribution of the source domain. Finally, the mapped target data is used with an autoencoder-based NBM pretrained on the source domain for anomaly detection. Presented results demonstrate more reliable anomaly detection despite target data scarcity. While this approach resembles domain mapping, it substantially deviates from its approaches. For one, it consists of only one-directional target-to-source mappings. Crucially, the mapping network is also unconstrained, i.e., there is no constraint to preserve specific WT states (e.g., maximal power generation or anomalous  behavior) to their corresponding state in the other domain. This may result in anomalous data being mapped to arbitrary healthy operation states. This lack of content preservation could therefore lead to unreliable fault detection. Our study overcomes such deficiencies by employing domain mapping based on CycleGAN with consistency losses to preserve the SCADA content during domain translation. Additionally, our work incorporates multi-hour long SCADA time series, as opposed to features from only a single time point.

\subsection{Domain mapping for unsupervised anomaly detection}
Few studies exist on domain mapping applications for unsupervised anomaly detection, even beyond the wind energy research field. Hardly any domain adaptation studies exist for anomaly detection tasks as in the present study where only normal training samples are available combined with limited data in the target domain \cite{yangAnomalyDetectionDomain2023}. To our knowledge, our study is the first to propose a CycleGAN-based approach to unsupervised anomaly detection for WT fault detection using domain mapping. 

Moreover, only few works have even performed the underlying domain translation with time series data, i.e., mapping time series, regardless of the task and outside the scope of wind turbine SCADA data. CycleGAN is proposed for data augmentation by generating artificial damaged states of acceleration data in \cite{luleciCycleGANUndamagedtodamagedDomain2023a} through a mapping of undamaged-to-damaged conditions. Pattnaik et al. \cite{pattnaikCycleGANBasedUnsupervised2022} translates time series across machines for bearing fault diagnosis using a CycleGAN framework. Unlike our anomaly detection study with scarce target data, their model maps abundant but unlabeled target domain data to the source domain where it is subsequently used in a fault classifier pretrained on the source domain. Results showed a significant outperformance compared to conventional domain adaptation methods when the domain shift is large, i.e., when mapping from substantially different but related machines.

%%%% --- DATASET ----- 
%%%% --- DATASET ----- 
%%%% --- DATASET ----- 
%%%% --- DATASET ----- 
\newpage
\section{Dataset}
\label{sec:dataset}

\begin{figure}[h!]
    \centering
    \includegraphics[width=1\textwidth]{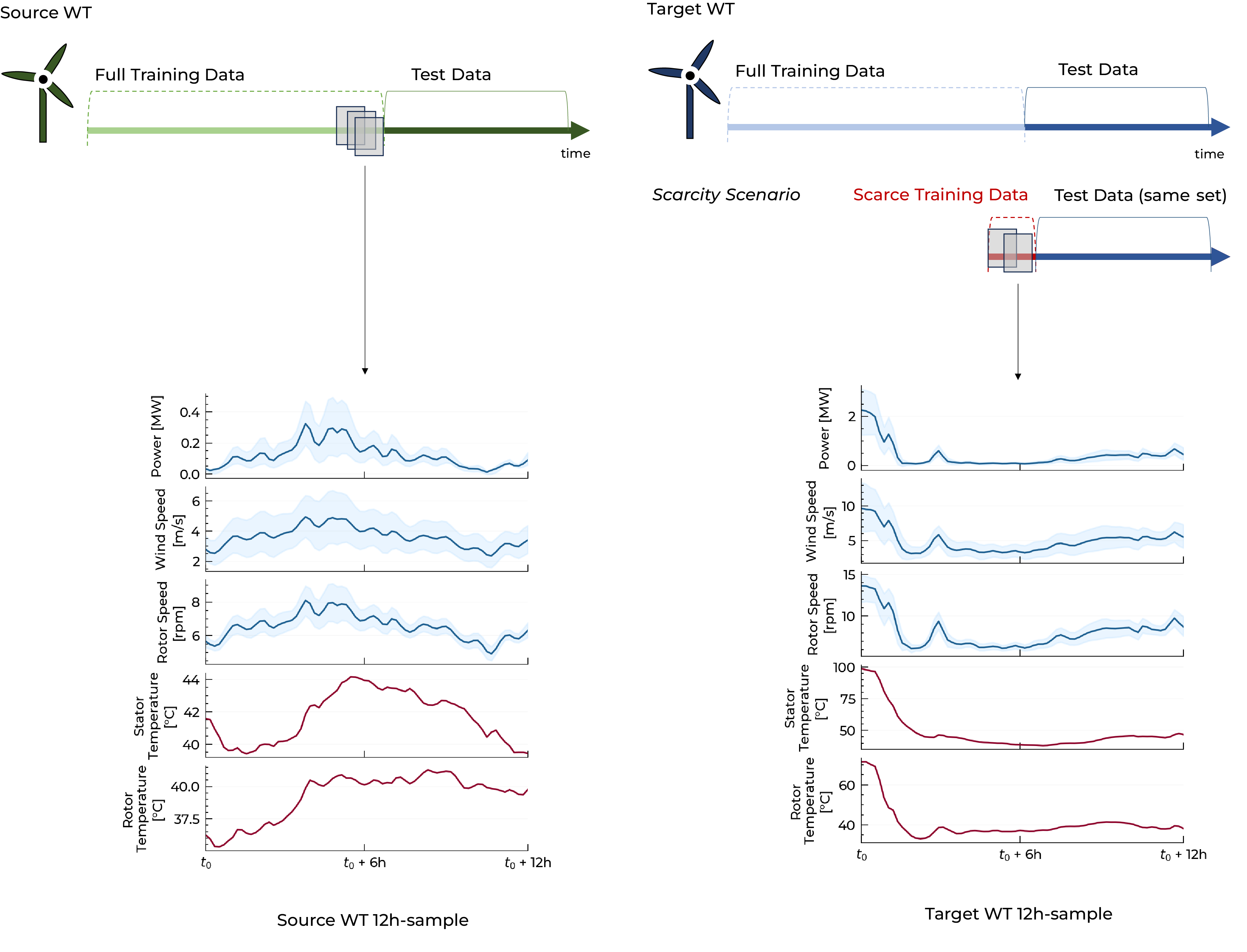}
    \caption{On the left, a source WT is selected with its data split into a training/validation and test set. On the right, we select a target WT which is also split into a preliminary training/validation and test set. For the target WT, a data scarcity scenario is then applied, in which only 1-8 weeks immediately preceding the target test set (which remains the same) are considered as the new training/validation set. For both WTs, we apply a sliding window approach to extract 12-hour SCADA samples from the sets, consisting of 11 channels in total. The shaded areas represent the corresponding maximum and minimum feature values.}
    \label{fig:scada}
\end{figure}

Our available raw data comprises 10-minute averaged SCADA data of 7 wind turbines, each from a distinct wind farm. While the turbines share the same manufacturer to ensure a matching SCADA variable system, they exhibit significantly different characteristics, model types, power rating, maintenance and fault history, and geographical location. An incident log is further provided for each WT. These logs contain a binary flag for each 10-minute measurement, describing time frames during which unknown and unspecified incidents occurred in the WT operation. These incidents do not necessarily represent anomalous behavior or faults. Thus, we do not use incidents for evaluation or classification but only for additional filtering, as we consider no measurements during these marked time frames as coming from normal operation. A more detailed overview and an illustration of the 7 WTs is provided in Appendix D.

Our study considers source-target pairs of WTs for domain adaptation, with one data-rich WT defining the source domain and the pair's second data-scarce WT defining the target domain. Our goal is to map SCADA samples between the source domain, consisting of a randomly selected WT, and the target domain, represented by another WT of a different WT type from a different wind farm. The data from the source domain WT is split into a training set (first 70\% of the data), a validation set (30\% of the training data), and a test set (the last 30\% of data). For the target domain WT, starting with its full dataset, we first apply the same 70-30-30 split as with the source WT. As a next step we apply a data scarcity scenario, by artificially shortening the training and validation set. Specifically, we shorten the training/validation set to consecutive time frames of 1 to 8 weeks immediately preceding the test set, which remains the same across all scarcity scenarios. To ensure dataset size consistency when comparing across different WTs, we always refer to one week of data to comprise $1008$ (6 10-minute sample per hour * 24 * 7 days) SCADA samples instead of selecting by calendar week, as maintenance, data gaps, or incidents could cause strongly changing dataset sizes across turbines.  

These datasets are further normalized according to statistics calculated over the training set of the source domain such that each SCADA variable falls into a value range of $[-1,1]$. The target domain data is normalized using the same min-max normalization formula but using source domain statistics and therefore not confined within this range.

From each set, we apply a sliding window approach to extract a sample dataset consisting of 12-hour samples of consecutive SCADA measurements with 11 variables. Selected were the mean, maximum, and minimum wind speed, rotor speed, power output values, as well as two temperatures from internal components, namely the mean stator and rotor temperature. Due to the one-to-one mapping inherent with CycleGAN \cite{manytomanyGAN}, we refrained from using variables with a larger stochastic component such as the ambient temperature. For instance, an idle WT operational state, represented through our highly correlated 11 input features, could exist with numerous possible ambient temperature variations, which may necessitate probabilistic domain mapping (many-to-many mappings, e.g.  \cite{manytomanyGAN}). The chosen 12-hours time frame allows us to better capture temporal dynamics within a sample, for instance thermal processes. An illustration of our dataset splits and resulting SCADA samples is shown in Figure \ref{fig:scada}.

Normal behavior models require that only normal samples, i.e., only healthy WT operation states, are used for training. We apply WT-specific filtering to exclude possibly abnormal samples from the training and validation sets. First, we exclude all measurements that fall into a time frame marked in the incident logs. Moreover, we apply a rated power filter rule and Mahalanobis distance-based filtering based on \cite{mckinnonComparisonNovelSCADA2022} to remove outliers and measurements from curtailment. No filtering is performed on the test sets.

%%%% --- METHODOLOGY ----- 
%%%% --- METHODOLOGY ----- 
%%%% --- METHODOLOGY ----- 
%%%% --- METHODOLOGY ----- 
%%%% --- METHODOLOGY ----- 

\section{Methodology}
\begin{figure}[h!]
    \centering
    \includegraphics[width=.7\textwidth]{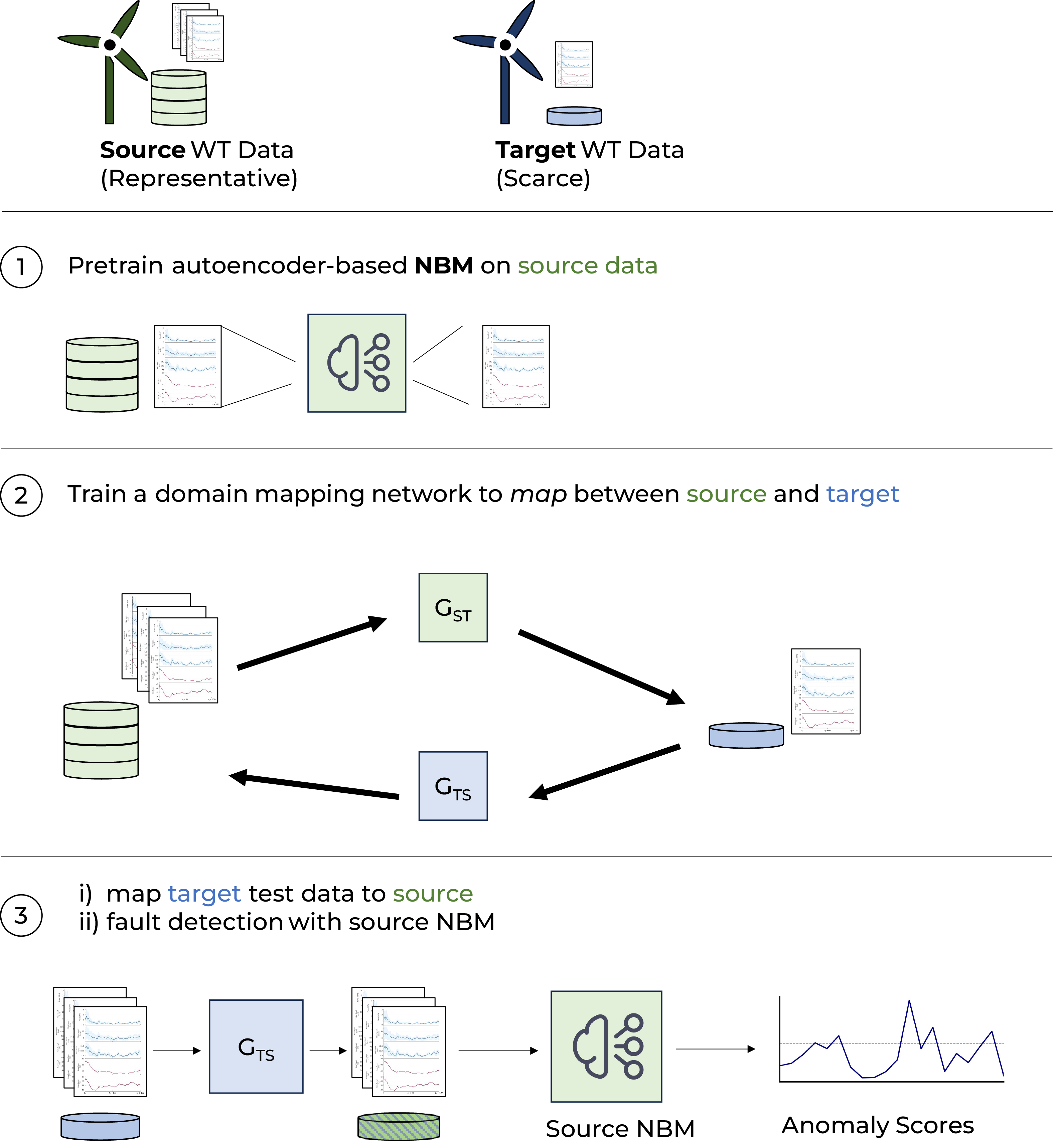}
    \caption{Workflow of the proposed approach. A source domain, represented by a WT with abundant data, and a target domain, represented by a WT with scarce data, are available. In step (1), an NBM is trained on the healthy, abundant source data only. This NBM can detect anomalies for the source WT but cannot be used for the target WT due to the domain shift. In (2), our proposed domain mapping technique using a CycleGAN is applied. The network learns to map source data to match target data and vice versa. In final step (3), the test set data of the target WT is first mapped to the source domain using the trained target-to-source network component ($G_{TS}$). As it now resembles source data, the pretrained NBM can be applied to obtain anomaly scores on the mapped data.}
    \label{fig:workflow}
\end{figure}

Our study aims to improve the reliability of fault detection models for WTs with unrepresentative training data by leveraging models and data from similar, but not identical, WTs for which representative training data is available. To achieve this, we employ a generative domain adaptation approach, namely domain mapping. This section outlines our machine learning models, domain mapping losses, and evaluation metrics used. Our proposed framework consists of a domain mapping network that can translate source domain samples (consecutive 12-hour SCADA samples) into the target domain and vice versa. Despite the translation being learned cyclically and in both directions during training, we are ultimately interested in mapping data from the data-scarce target WT to the data-rich source WT for subsequent anomaly detection. Once the data is mapped, the anomaly detection is achieved through evaluations with an autoencoder-based NBM pretrained on the source domain. The proposed workflow is shown in Figure \ref{fig:workflow}.

\subsection{Autoencoder-based NBM}
We employ an NBM based on autoencoders \cite{liComprehensiveSurveyDesign2023} for anomaly detection and fault detection. Autoencoders are unsupervised neural networks trained to encode input data into a lower dimensional representation (encoder) and to subsequently reconstruct the input data from the compressed representation (decoder). To reconstruct the input as exactly as possible, an autoencoder learns to encode the most important features, thereby learning critical variable and feature relationships. An input sample that significantly deviates from the training set (i.e., an anomaly) will thus result at test time in an elevated reconstruction error, representing an anomaly score. Autoencoders have been employed as WT-specific NBMs (e.g., \cite{jonasVibrationFaultDetection2023, roelofsAutoencoderbasedAnomalyRoot2021, renstromSystemwideAnomalyDetection2020}).
Our autoencoder takes as input 12-hour SCADA samples with 11 channels (792 data points) and learns to reconstruct them based on an encoding size (bottleneck dimension) of 72 units. A model is trained on only normal source domain training samples for each WT domain pair. 
At test time, we categorize all samples with a reconstruction error (mean absolute error between the autoencoder reconstruction output and its original input) above the defined WT-specific threshold as anomalous. The threshold is set to detect far out outliers \cite{schwertmanSimpleMoreGeneral2004} for all models as $T=q_3+3 \left(q_3-q_1\right)$, where $q_3$ and $q_1$ represent the 75th and 25th percentile of the normal validation data reconstruction errors, respectively. Further information about the autoencoder model is outlined in Appendix A. 
\newpage

\subsection{Domain mapping model}
\begin{figure}[h!]
    \centering
    \includegraphics[width=.85\textwidth]{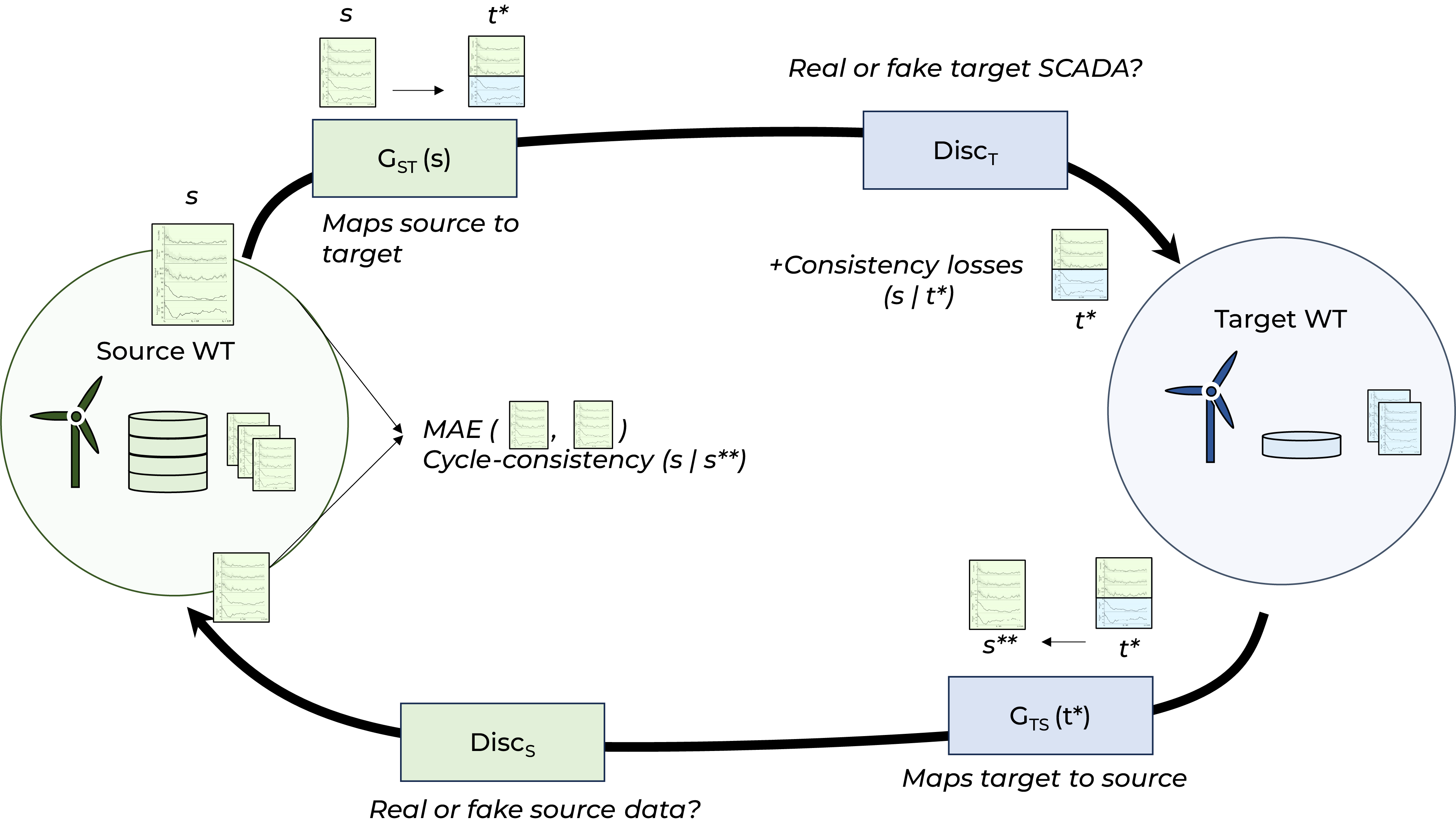}
    \caption{Illustration of the proposed domain mapping network. Visualized here is only the translation of a SCADA sample from the source domain ($s$) to the target domain and back, while our model maps in both directions. A source sample ($s$) is mapped by the generator for the respective domain direction ($G_{ST}$) to resemble data from the target domain under additional content-preservation constraints. The mapped sample ($t^\ast$) is mapped back to the source domain using $G_{TS}$, completing the cycle. The cycle-consistency forces the cycled sample ($s^{\ast\ast}$) to closely resemble the original input.}
    \label{fig:domainmappingnetwork}
\end{figure}

Our domain mapping network is based on the CycleGAN formulation \cite{zhuUnpairedImagetoImageTranslation2017}. The network consists of two one-directional generative adversarial networks (section \ref{sec:models}) taking as input SCADA samples of one WT to transform them into data resembling the other WT. The generators are trained to synthesize realistic samples in an adversarial way by trying to outperform the WT-specific critics, which try to distinguish between fake (generated) and real samples of their particular WT. Unconstrained generators, as in the case of \cite{jinConditionMonitoringWind2023}, can however potentially transform an input SCADA sample in any way and disregard the inherent content within the sample, which represents a particular state of the WT, e.g., idle power, maximal power production, or anomalous behavior. We propose and demonstrate that adding consistency losses (section \ref{sec:contentconsistency}) to the network enforces the content-preservation for anomaly detection. Most importantly, a cycle-consistency loss enforces that a sample first mapped to the other domain and then back to its original domain should remain similar. Figure \ref{fig:domainmappingnetwork} illustrates the concept of our domain mapping network.

In the following, we outline the specifications of the generators, discriminators, and content-preserving losses. Formally, we define a source domain $\mathcal{D}_\mathcal{S}$ representing the source WT and a target domain $\mathcal{D}_\mathcal{T}$ representing the target domain WT with domain datasets $D_S,D_T$ containing respective SCADA samples here labeled as $s$ and $t$.

\subsection{Generators and critics}
\label{sec:models}
The framework consists of two generators $G_{ST},G_{TS}$ and two discriminators ${Disc}_T,{Disc}_S$. The generators are tasked to map a sample from one domain, which is their input, to resemble a sample from the other domain. There is one generator for each domain direction, i.e.,  $G_{ST}:\mathcal{D}_\mathcal{S}\rightarrow\mathcal{D}_\mathcal{T},G_{TS}:\mathcal{D}_\mathcal{T}\rightarrow\mathcal{D}_\mathcal{S}$. The generator $G_{ST}$ maps an input SCADA sample from $\mathcal{D}_\mathcal{S}$ to resemble a sample belonging to domain $\mathcal{D}_\mathcal{T}$. As opposed to the standard GAN formulation, the generators take as inputs real samples of a domain to translate, instead of synthesizing samples from random noise. The discriminators generally act as critics, assessing whether a sample belongs to the underlying probability distribution or not. ${Disc}_T\left(x\right)$ is generally a score reflecting whether a sample candidate $x$ belongs to the corresponding domain distribution $P_T$ or not. In our work, we use the GAN-QP framework \cite{suGANQPNovelGAN2018, liSystematicSurveyRegularization2023} as in \cite{jinConditionMonitoringWind2023} to train the generators and critics to generate realistic samples. Further details about model architectures, hyperparameter optimization, and complete training procedure are presented in Appendix B.

\subsection{Content preservation losses}
\label{sec:contentconsistency}
As the goal is to ultimately perform fault detection with mapped data, it is critical to ensure that the generator mapping preserves the SCADA sample content, i.e., the operational state represented by the sample such as constant maximum power output and in particular anomalous behavior. The proposed cycle-consistency loss in CycleGAN \cite{zhuUnpairedImagetoImageTranslation2017} constrains the domain mapping to encourage that a mapped sample mapped back to its original domain resembles the original sample. This constraint aims to ensure that during a cycled mapping there were no changes or features added or removed. As an example, a zebra mapped to a horse and then back should remain the same animal in the same position and field, as otherwise content alterations may have been performed by the models. Formally, it encourages $G_{ST}\left(G_{TS}\left(t\right)\right)\approx t$ and vice versa. This is achieved by adding an L1 loss to the generator loss punishing deviations between original and cycled samples:

\begin{equation}
    \mathcal{L}_{cyc}=\lambda_{cyc}\ (MAE\ (t,\ G_{ST}\left(G_{TS}\left(t\right)\right))\ +\ MAE\ (s,\ G_{TS}\left(G_{ST}\left(s\right)\right)))
\end{equation}

However, the cycle-consistency loss can be in our case insufficient to ensure a consistent content mapping across domains. To illustrate this, let us consider adding a hypothetical flipping operation to the generators. Typically, with image data, flipping an input will cause discriminators to reject the mapping. For instance, flipping a horse image upside down before translating it into a zebra will never generate a realistic zebra, whereas “flipping” a SCADA input (e.g., maximum power mapped to zero power) before translation can still generate realistic domain samples. If the same flipping operation is performed in the opposite domain direction, the cycled output will resemble the original input, thereby still enforcing the cycle-consistency but without actually preserving content. To account for this, we further restrict the mapping space with additional consistency losses. 

We add two physics-informed loss functions to construct our content-consistency loss. The first loss encourages the generators to map idle states in the original domain, for instance, zero active power output or no rotor rotation, to idle states in the other domain. Formally, we define all positions of selected channels (in our study: minimum, mean, maximum power and rotor rotation) within a SCADA sample $x$ as a zero state if they are zero valued: $Z:=x_{c,i}=0; i=1,...,72$, where $c$ denotes the channel. The zero loss discourages deviations from the zero state when mapping to a domain:

\begin{equation}
    \mathcal{L}_{0}=\lambda_{0}\ (MAE(G_{ST}(
    s\mathbbm{1}_{Z}), 0) + MAE(G_{TS}(
    t\mathbbm{1}_{Z}), 0))
\end{equation}

The rated power loss encourages that power outputs at a rated WT value remain at a rated value for the corresponding other WT. Let $C_S$ and $C_T$ represent the rated power of the source and target domain WT, respectively. All positions within a SCADA sample $x$ are defined to be at a rated power if they match the WT capacity: $R:=x_{c,i}=C_D; D=\text{domain (S, T)}, i=1,...,72$, where $c$ denotes the channel (in our study: the mean power). The rated power loss is then defined as:

\begin{equation}
    \mathcal{L}_\mathcal{R}=\lambda_\mathcal{R} (MAE(G_{ST}(s\mathbbm{1}_R),C_T) 
    + MAE(G_{TS}(t\mathbbm{1}_R),C_S))
\end{equation}

\begin{figure}[!htp]
    \centering
    \includegraphics[width=.9\textwidth]{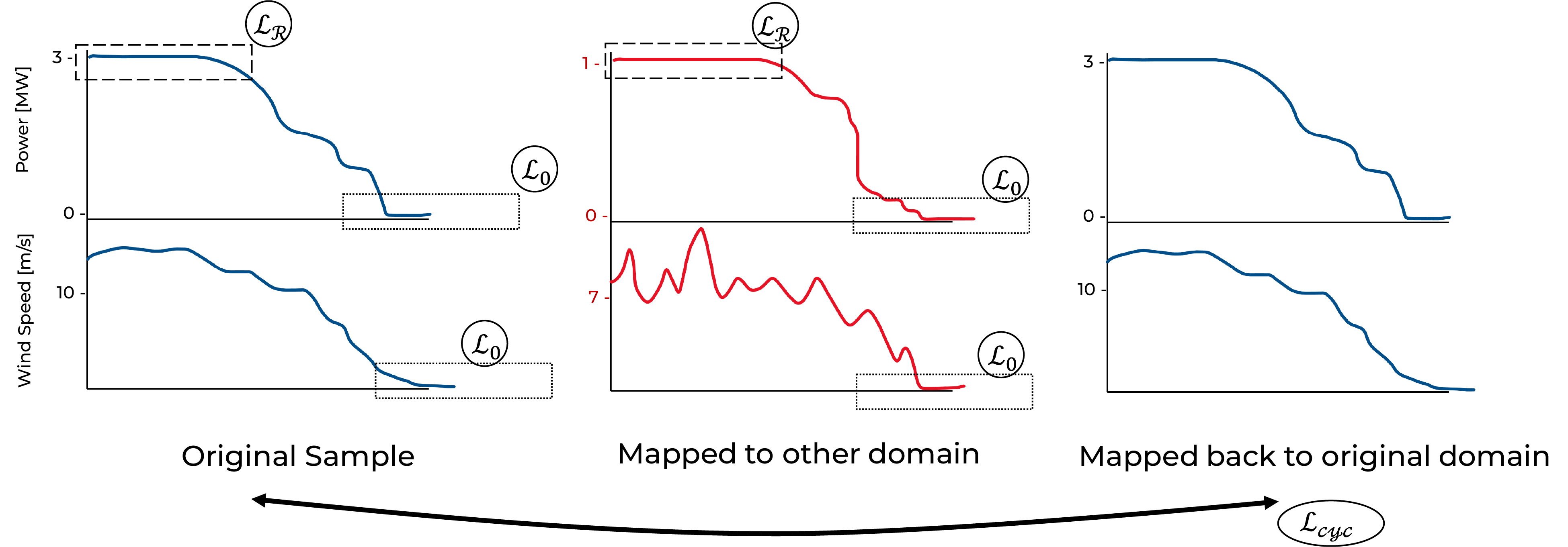}
    \caption{A sketch illustrating our content-consistency losses. An illustrative 12h-sample containing only the WT power output and wind speed is shown on the left, with its mapping to the other domain in the middle and mapped back to the original domain on the right. The zero-loss enforces that zero states (e.g., zero power) should be mapped to zero states in the other domain, shown by the boxed dashed regions marked by $\mathcal{L}_0$. The rated power loss ($\mathcal{L}_R$) enforces that when a WT is running at maximum capacity (in the figure: 3 MW) it should also run at the rated capacity in the other domain (in the figure: 1 MW). Finally, the cycle-consistency loss $\mathcal{L}_{cyc}$ ensures that the cycled sample (mapping a mapped sample back to its original domain) remains similar to its original sample.}
    \label{fig:consistencylosses}
\end{figure}

The consistency losses are added to the generator loss with relative weights $\lambda_{cyc}, \lambda_0, \lambda_R$, determined using a hyperparameter search. More details are outlined in Appendix B. We illustrate our consistency losses in Figure \ref{fig:consistencylosses}.

\subsection{Evaluation and benchmarks}
Ultimately, we are interested in obtaining similar anomaly scores with scarce target data as if we had abundant target training data available. Therefore, we consider the test set anomaly scores from an NBM trained on abundant target training data (i.e., without any data scarcity scenario applied) as our ground truth in this study. More specifically for anomaly detection, we are interested in whether compared anomaly scores both exceed their (model-specific) threshold or not (anomalous or normal). We introduce a similarity measure in the following to compare data scarce models to the representative NBM. 

Let $a^\ast=({a_1^\ast},\ldots,{a_n^\ast})$ be the ground truth test set anomaly scores of the NBM trained on the full target domain training data (i.e., no data scarcity scenario applied), consisting of $n$ test set samples with a model-specific threshold $T^\ast$. Let $a=(a_1,\ldots,a_n)$ be the test set anomaly scores of a compared NBM (e.g., anomaly scores of mapped data) with its threshold $T$. Anomaly scores are converted into a binary value expressing whether the score exceeds the threshold (positive, 1) or not (negative, 0), i.e.,  $y^\ast=a_1\geq T^\ast,\ldots,a_n\geq T^\ast$  and $y=a_1\geq T,\ldots,a_n\geq T$. We compare the binary values using classification metrics to obtain the performance as the F1-score, defined as:

\begin{equation}
    \text{F1-Score} = \frac{2\text{TP}}{2\text{TP}+\text{FP}+\text{FN}}
\end{equation}

\begin{figure}[!h]
    \centering
    \vspace{3ex}%
    \includegraphics[width=1.05\textwidth]{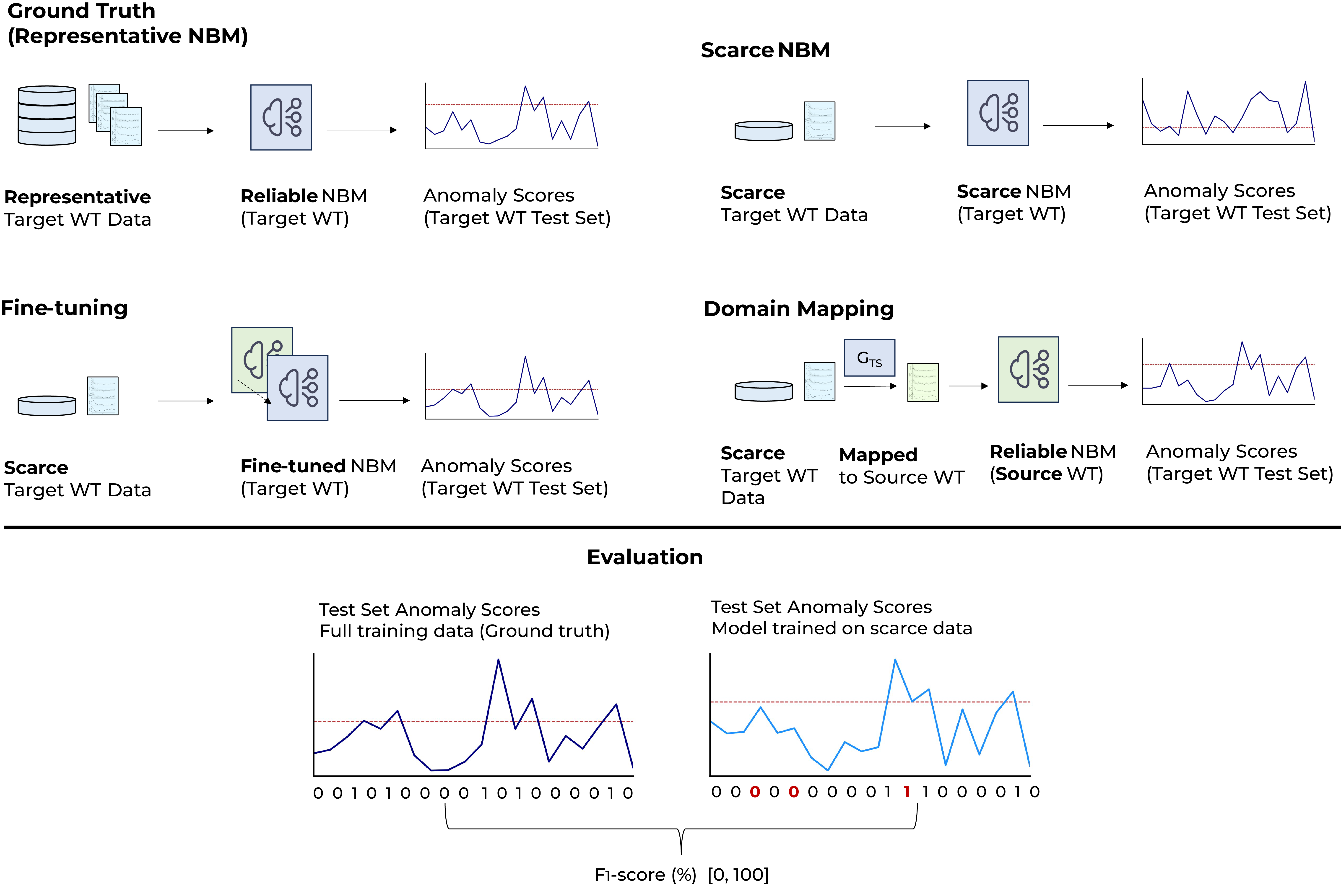}
    \caption{All approaches for comparison are illustrated at the top. In the upper left, we consider an NBM trained on abundant target WT data (no data scarcity) producing anomaly scores on a fixed test set, representing our ground truth. We compare these anomaly scores with an NBM trained on scarce target training data only (upper right), a fine-tuning approach (lower left) and our domain mapping network (lower right). These obtained anomaly scores, sharing the same test set, are compared in terms of a threshold similarity (F1-score) to the ground truth (illustrated at the bottom).}
    \label{fig:evaluation}
\end{figure}

We compare the performance of our domain mapping technique with two benchmarks: The first one is an NBM trained on scarce target data only, providing a baseline without domain adaptation and SCADA data from existing wind farms. The second one is a fine-tuning benchmark that represents the performance of a conventional and simple domain adaptation approach for this task. Starting with a pretrained NBM trained on SCADA samples of the source domain WT, we fine-tune the model on the available scarce target training data. Our benchmark models and the evaluation are illustrated in Figure \ref{fig:evaluation}.

%%%% --- RESULTS & DISCUSSION ----- 
%%%% --- RESULTS & DISCUSSION ----- 
%%%% --- RESULTS & DISCUSSION ----- 
%%%% --- RESULTS & DISCUSSION ----- 
%%%% --- RESULTS & DISCUSSION ----- 

\newpage
\section{Results and Discussion}

\subsection{Fault detection performance}
We assess the fault detection performance of our proposed domain mapping network in terms of producing similar anomaly scores as a target WT NBM trained on abundant data, as outlined in Figure \ref{fig:evaluation}. In total, we evaluated 28 source-target WT domain pairs using 6 distinct target domain WTs, as one of the seven WTs (WT07) was used during model selection and hyper-parameter optimization (see Appendix B), with varying degrees of target training data scarcity of 1-8 weeks. All domain pairs consist of a source and target WT differing in turbine specifications in terms of model and having a unique rated power capacity (see Appendix D), farm, and geographical location. For each source-target domain pair, we compared the performance to an NBM trained on scarce data only and a fine-tuning benchmark. 

\input{Tables/mean_f1_scores}

\paragraph{Training NBMs on scarce data.} For all 6 target WTs considered in our study, we first trained NBMs using only scarce target training data under varying scarcity scenarios. That is, we trained an NBM using 1-8 weeks of training data and evaluated the resulting anomaly scores on the fixed test set in comparison to the ground truth. Our findings in Table \ref{tab:mean_f1_scores} show that NBMs trained on limited training sets achieved comparably poor fault detection performance, with the mean F1-score generally decreasing as less training data becomes available. While a training set comprising 2 months of data achieved on average an F1-score of 84.3\% across all 6 target WTs, the mean score substantially decreased to 51.3\% when only 2 weeks are considered, showing a strong drop in similarity of anomaly scores compared to the NBM trained on a full training dataset. Limited data can lead to unrepresentative training sets with an insufficient coverage of samples from normal operational behavior. An example of representative and unrepresentative training data is shown in Appendix E. Thus, the autoencoder-based NBM becomes incapable of reconstructing normal data at test time that was missing from the training set. Our results indicating that training NBMs under data scarcity can result in unreliable fault detection support the findings of e.g., \cite{grataloupWindTurbineCondition2024, jenkelPrivacyPreservingFleetWideLearning2023}.

\paragraph{Improving fault detection with domain adaptation.} To overcome these limitations, we employed domain adaptation techniques. We trained and evaluated our proposed domain mapping technique and fine-tuning as comparison benchmark to overcome these limitations. We present the achieved improvements in F1-score compared to training an NBM on scarce data only (i.e., baseline without domain adaptation) for each domain pair in Table \ref{tab:f1_score_diffs}, further averaged by data scarcity degree. The absolute F1-scores are available in Table \ref{tab:f1tablefull} in Appendix C.  Both fine-tuning and our domain mapping network achieved an improvement in fault detection performance in scenarios with strong data scarcity, i.e., particularly in situations where the scarce NBM fails to achieve reliable fault detection.

\input{Tables/f1_score_diffs}
 As shown in Table \ref{tab:f1_score_diffs}, both domain adaptation methods managed to to improve fault detection performance over training on scarce data alone in most domain pairs and across almost all data scarcity degrees. Fine-tuning was able to improve the performance by increasing the F1-score on average by +4.7\% points when only 1 month of target training data was available and +9.3\% with 2 weeks of data, showing improvements across all scarcity scenarios except 2 months. These results further suggest that fine-tuning can be employed to adapt knowledge embedded in the source domain NBM to a limited target domain. 
 
 Notably, our proposed domain mapping approach outperformed fine-tuning across all considered training set sizes, especially in scenarios with severe data scarcity. In the scenarios ranging from 1 week to 1 month of data, our domain mapping approach achieved a substantial performance gain over not only the NBM trained on scarce data but also consistently over fine-tuning, with an increase in mean F1 score by +10.3\% points for 1 month and +16.8\% for 2 weeks. Our presented domain mapping approach makes use of the entire source WT's dataset to learn mappings and uses its reliable model for anomaly detection, as opposed to fine-tuning, which exclusively relies on only the trained source NBM parameters being adjusted by a limited target training set, thereby limiting its acquired knowledge of the domain shift. Our findings highlight the effectiveness and potential of our domain mapping method as an alternative approach to conventional fine-tuning to mitigate the challenges of data scarcity in WT fault detection. Our proposed technique can enable more reliable and earlier fault detection, for instance for newly installed wind turbines, by incorporating models and data of only one reference source wind turbine with abundant data. 

\paragraph{Performance decreases with abundant data.} Conversely however, in scenarios with abundant data (e.g., 2 months), both fine-tuning and domain mapping showed limited improvements or even significant performance drops. A large decrease in performance can particularly be observed when the scarce NBM already achieved comparably high F1-scores, indicating representative training data (e.g., target WT04 in Table \ref{tab:mean_f1_scores}). This suggests that when enough data is available to train a reliable NBM, the benefits of these domain adaptation techniques may diminish. A drop in performance can likely be attributed to a loss of information through the fine-tuning and domain mapping operation, coupled with our hyperparameter optimization for both methods having been performed for 1 month of available data, which may result in suboptimal training when abundant data is available. That is, our domain adaptation methods were optimized to be more constrained. In principle, fine-tuning should be able to overfit on the target data (completely disregarding any source knowledge), thereby achieving the same performance as if it were a new model exclusively trained on target data. Equivalently, the content-consistency loss constraints of our domain mapping could be more relaxed, allowing our model to generate samples more freely when enough diverse samples are available. Such adjustments would however require a priori information about the representativeness of the training set compared to the test set (which is by definition unavailable at training time), consequently restricting the possibilities for adequate case-by-case tuning. More research is needed to identify and counteract the loss of information as well as to improve hyperparameter optimization to preemptively detect or mitigate performance decreases in these scenarios. 

\paragraph{WT-specific differences in performance changes.} Lastly, we note strong differences in performance across target WTs. These could be attributed to the randomness of the selected training set weeks (e.g., weather conditions, variety of operational states), test set characteristics (e.g., number of faults), or general turbine behavior. Further research is needed to investigate these variations and to possibly adjust the network and hyperparameters to the specific characteristics of WTs. Furthermore, our findings show significant variations across WT pairs. A possible cause of performance differences may be, among other, that WT pairs can exhibit varying domain shifts. Thus, selecting suitable source-target pairs can be important. However, the fundamental unavailability of representative validation data impedes an informed pair selection. An exploration is needed to investigate and assess the suitability of WT pairs for domain mapping, which may involve e.g., a quantification of the domain shift. For certain domain pairs, fine-tuning may also retrospectively turn out to be a better solution compared to domain mapping. In some few other cases, both domain adaptation methods may even lead to a substantial drop in performance even with limited training data. Further investigations are required to clearly identify these causes and for finding mitigation strategies. These variations may cause uncertainties regarding an appropriate method selection, as due to the inherent lack of knowledge about the test set, it remains unclear which approach to choose at training time.

%% case study 
%% case study 

\subsection{Detailed analysis of a domain pair example}
\label{sec:example}
We present a detailed comparison of the fault detection performance across all methods for one specific domain pair with the source domain WT “WT07” and target domain WT “WT05” with 1 month of available training data. The performance gains for this domain pair represent approximately the average for this scarcity scenario. 

\subsubsection{NBM trained on representative data}
\begin{figure}[ht!]
    \centering
    \includegraphics[width=.65\textwidth]{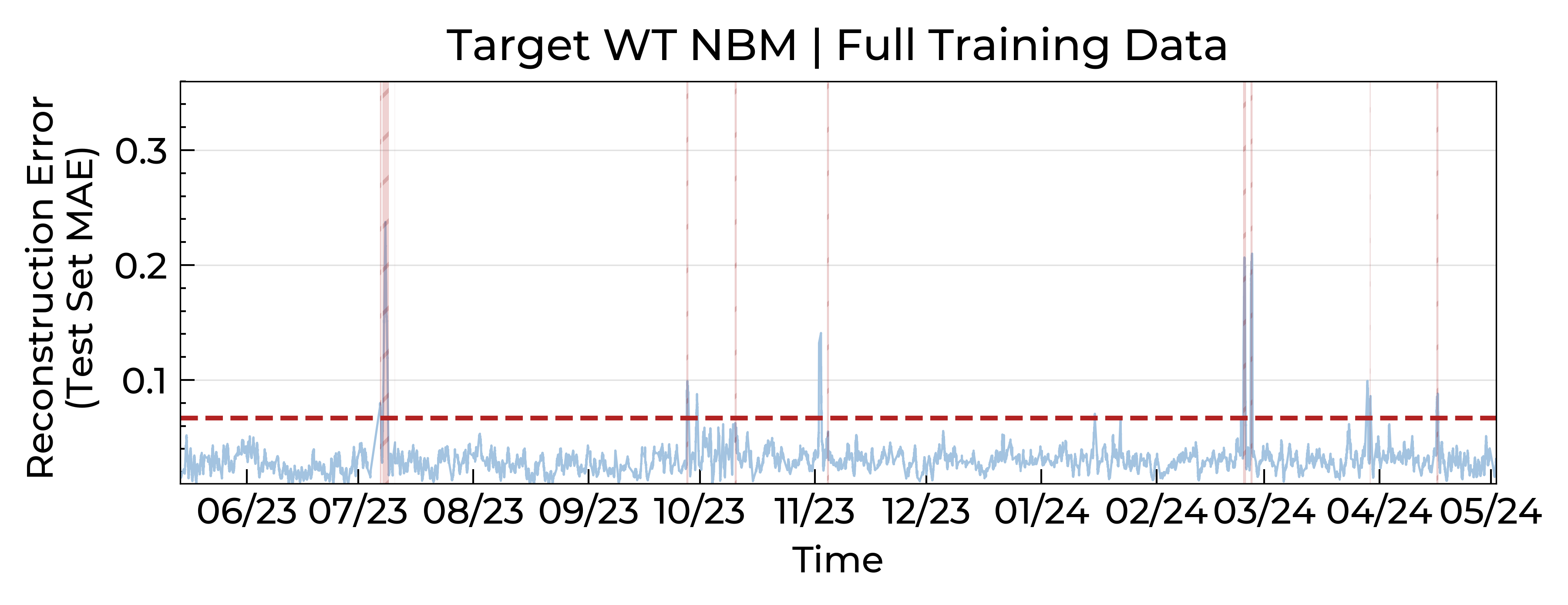}
    \caption{Anomaly scores, i.e., reconstruction errors, of the target WT test set obtained from the representative NBM. The model-specific threshold is shown in the horizontal red dotted line, while time frames with logged incidents are shown by red shaded areas.}
    \label{fig:fullNBM}
\end{figure}

To obtain our ground truth anomaly scores for the target WT test set, we trained an NBM on the full target training data, which contains measurements recorded over a span of more than 2 years (Appendix D). We visualize the resulting anomaly scores, i.e., the NBM reconstruction errors, for the target WT test set in Figure \ref{fig:fullNBM}. The scores generally exceed the threshold during time frames when incidents were logged, yet not for all incidents (e.g., in mid-October 2023). As the incidents are unspecified, we are unable to categorize this as a missed fault or correct fault detection. During normal operation, the scores tend to remain below the threshold, although certain short periods exhibit elevated scores (e.g.,in the middle of January 2024), which we cannot determine as false positives or identified but unlogged abnormal states. However, we only focus on comparing these scores to the ones obtained trained with scarce data.

\subsubsection{NBM trained on scarce data}
\begin{figure}[htp!]
    \centering
    \includegraphics[width=.65\textwidth]{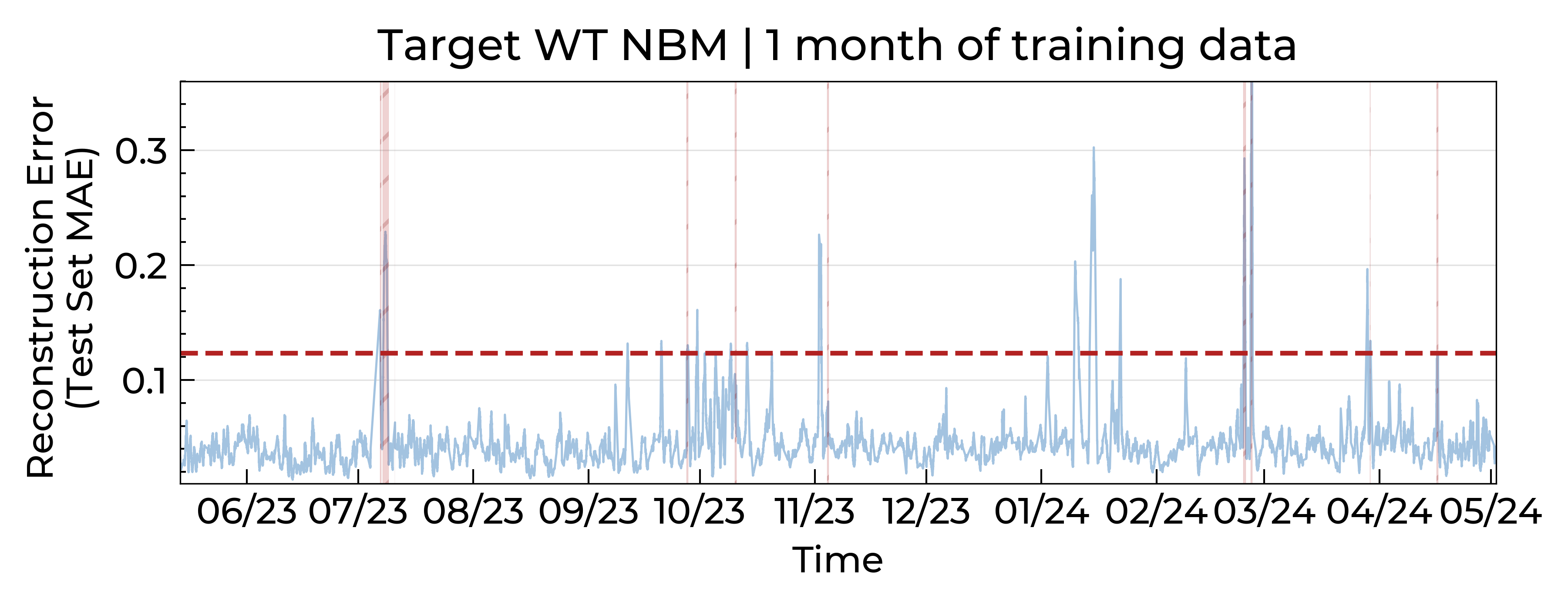}
    \caption{Anomaly scores of the always identical target WT test set obtained from the NBM trained on scarce training data.}
    \label{fig:scarceNBM}
\end{figure}

An NBM was then trained on only 1 month of target training data preceding the identical test set. The anomaly scores of this model are visualized in Figure \ref{fig:scarceNBM}. When comparing the threshold similarity, this model achieved an F1-score of 63.9\% (Table \ref{tab:f1tablefull}). Notably, numerous anomaly scores falsely exceed the threshold compared to their counterpart in the ground truth NBM (e.g., January 2024), while scores during incidents largely match the ones from the representative NBM. As autoencoders learn normal behavior from the training set, this could indicate that normal operational states were missing in the limited training set, causing high reconstruction errors at test time and therefore falsely elevated anomaly scores. Overall, a significant decrease in fault detection performance can be observed when only scarce training data is available.

\subsubsection{Fine-tuning the source domain NBM}
\begin{figure}[htp!]
    \centering
    \includegraphics[width=.65\textwidth]{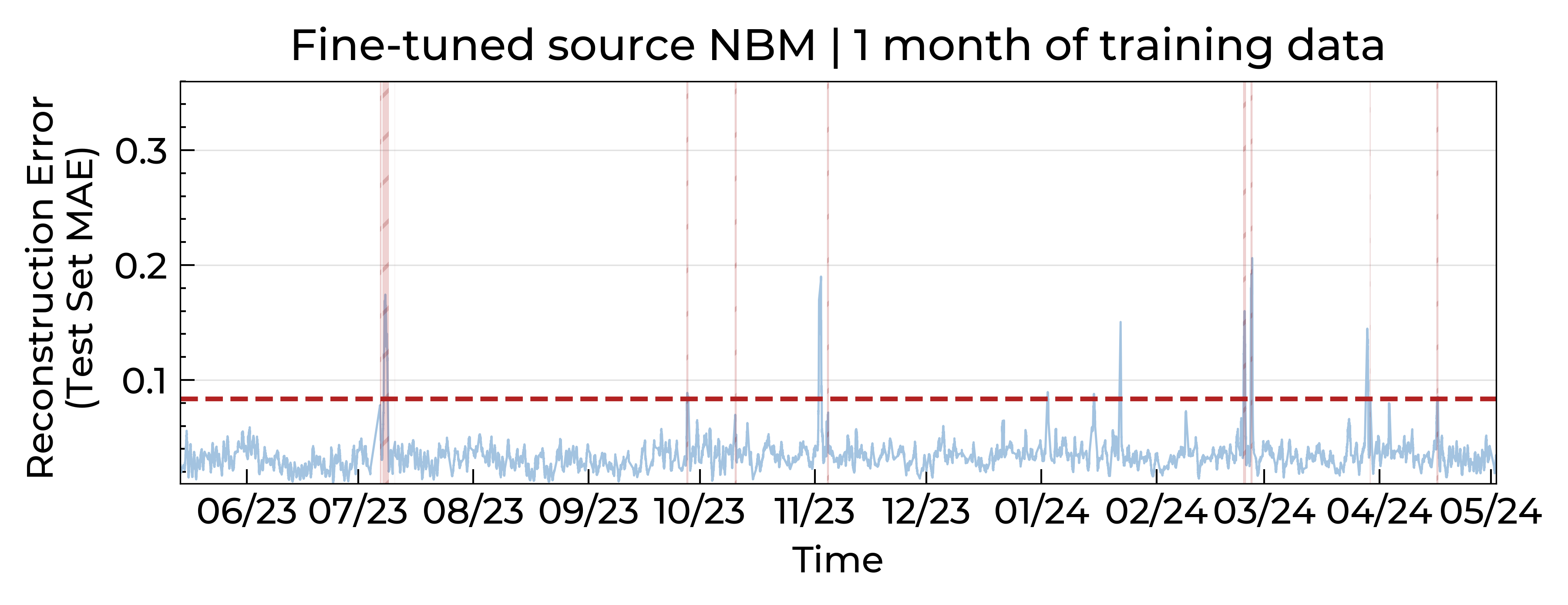}
    \caption{Anomaly scores of the target WT test set from the fine-tuned source NBM using scarce target training data.}
    \label{fig:FT_NBM}
\end{figure}

As a domain adaptation benchmark, an NBM was first trained on the abundant \textit{source} training data, i.e., the full training set of WT07, and then \textit{fine-tuned} using 1 month of target domain training data. Results are shown in Figure \ref{fig:FT_NBM}. This model achieved an increase in F1-score to 68.8\% (+4.9 percentage points) compared to the scarce NBM. In particular, we notice a decrease in falsely elevated scores (for instance, October 2023) while anomaly scores generally remain elevated when they are in the ground truth. Our results suggest that fine-tuning is capable of adjusting information embedded in the source domain NBM to the target domain, resulting in unseen normal data missing in the target WT to be still considered normal, enabling more reliable fault detection when training data is lacking.

\subsubsection{Domain mapping}
\begin{figure}[h!]
    \centering
    \includegraphics[width=.65\textwidth]{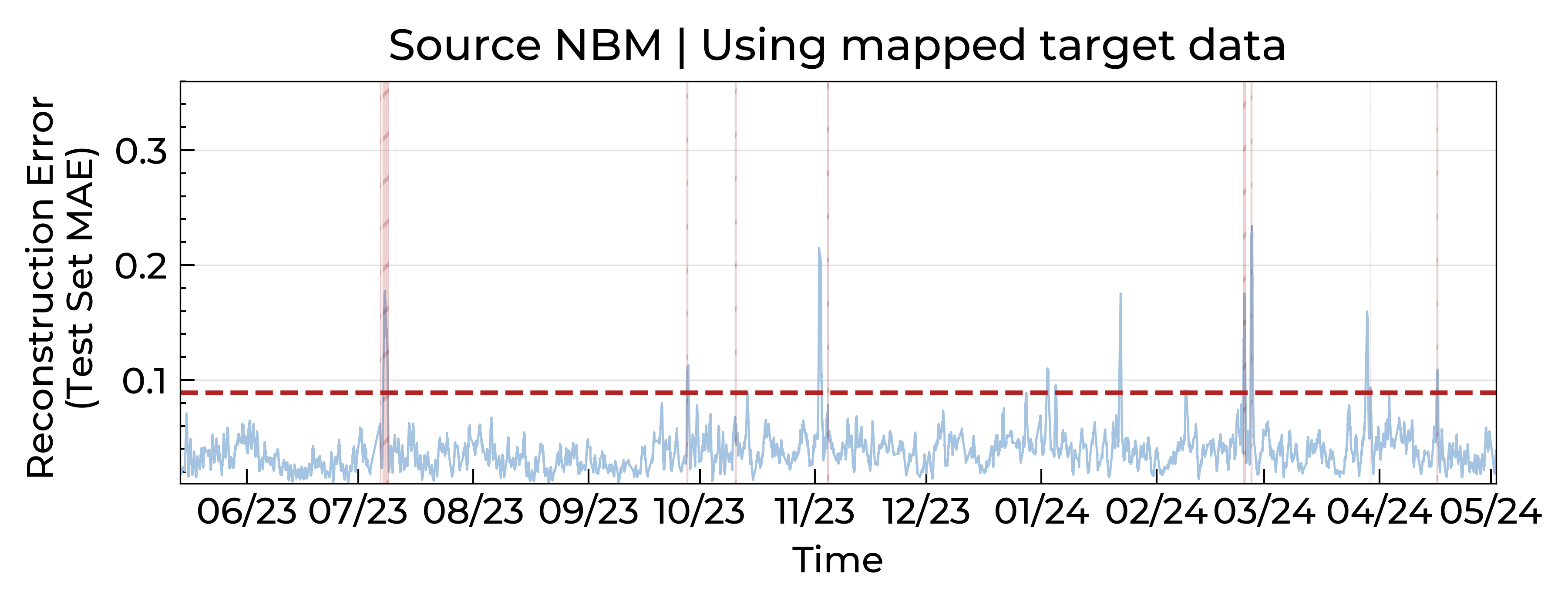}
    \caption{Anomaly scores of the target WT test set. The reconstruction errors are obtained from the pretrained source WT NBM using the mapped target test set data.}
    \label{fig:CG_NBM}
\end{figure}

A mapping network was trained to map samples from the source WT to the scarce target WT and vice versa. After training was finished, the target-to-source model ($G_{TS}$) was used to map the target WT test set samples to the source domain. The mapped samples were subsequently used with the source domain NBM to obtain anomaly scores, which are shown in Figure \ref{fig:CG_NBM}. Our proposed technique achieved further performance gains over fine-tuning. The F1-score was significantly increased to 76.5\% (+12.6), achieving the most similar anomaly scores compared to the ground truth.

\begin{figure}[h!]
    \centering
    \includegraphics[width=.725\textwidth]{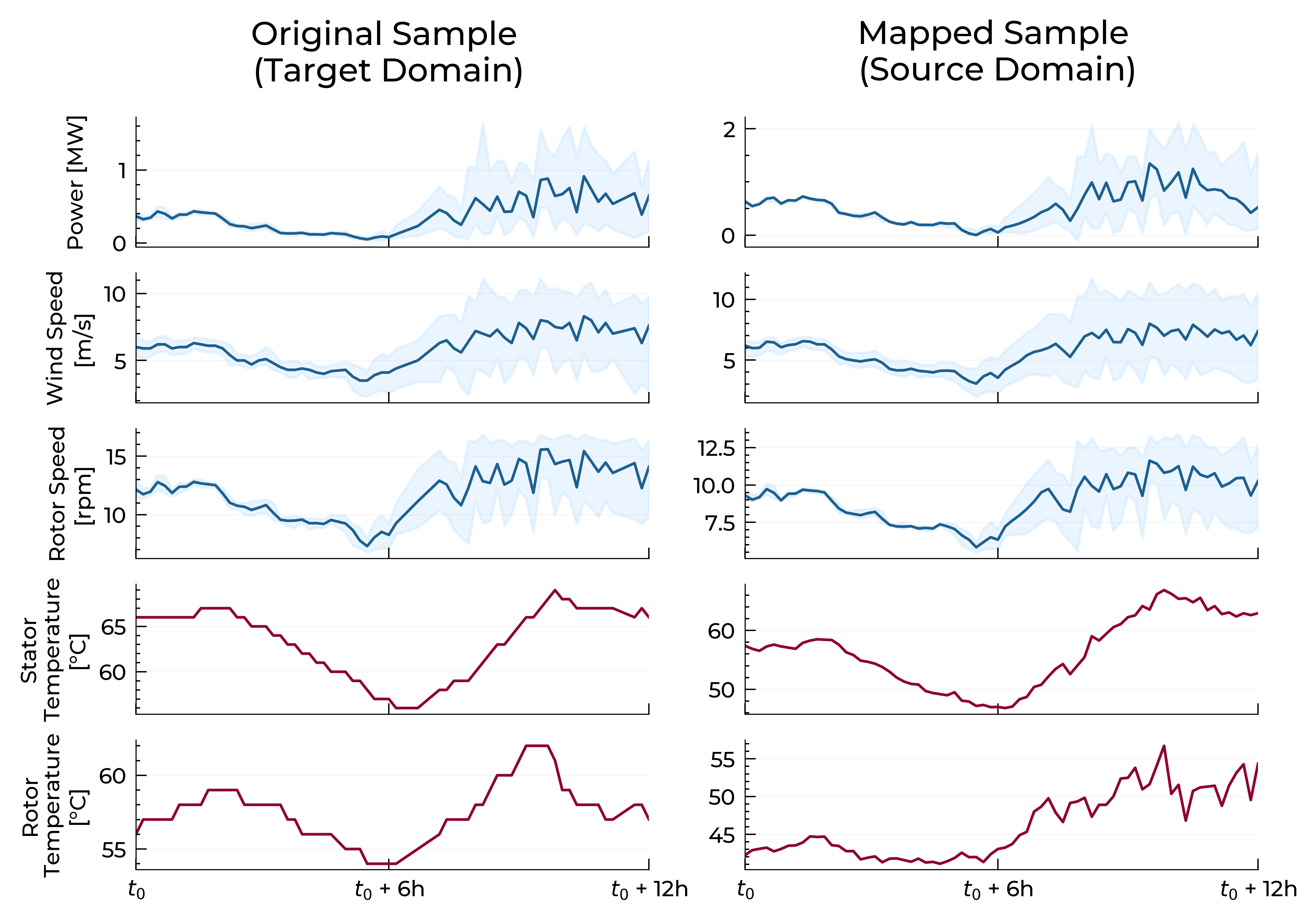}
    \caption{Example of the domain mapping result. A sample from the target WT's test set (left) is mapped by the trained network $G_{TS}$ to the source domain, shown on the right.}
    \label{fig:mapping_example}
\end{figure}

\newpage
This strong correspondence in anomaly scores indicates that our mapping network generates mappings closely resembling the source WT, as reconstruction errors for normal samples remain below the threshold, while additionally preserving the content, as anomalies are mapped to anomalous states. An example of a mapping using a target WT test set sample is illustrated in Figure \ref{fig:mapping_example}, showing preserved operational behavior, i.e., high and low power was mapped to a correspondingly scaled high and low power output with consistent component temperature behavior. 

% - - - -- - - -- - - -- CONCLUSION 
% - - - -- - - -- - - -- CONCLUSION 
% - - - -- - - -- - - -- CONCLUSION 
% - - - -- - - -- - - -- CONCLUSION 
% - - - -- - - -- - - -- CONCLUSION 

\section{Conclusion}
This study investigated the application of generative domain adaptation, specifically domain mapping, to address the challenge of limited training data when using normal behavior models in wind turbine fault detection. NBMs require substantial normal operation data, which when missing can result in unreliable models. To address this challenge, we proposed a novel domain adaptation approach that leverages data from a data-rich source domain WT of a different wind farm. Our domain mapping technique transforms SCADA data from a data-scarce target domain WT to resemble data from a different WT with representative training data. This mapping allows for the use of the source WT's NBM for fault detection in the target WT despite limited data. We validated our method by conducting experiments using 28 different WT pairs across varying degrees of data scarcity (1 to 8 weeks). Our findings demonstrate the superior performance of generative domain adaptation compared to training models on limited data and conventional fine-tuning, particularly when faced with severe data scarcity. Our results highlight the significant potential of domain mapping in achieving earlier and more reliable WT fault detection models when only scarce training data is available, for instance in newly installed wind farms.

Our study has several limitations which provide opportunities for future research. First, further exploration of time-series-based model architectures and frameworks, consistency losses, and generalizable hyperparameter optimization would be useful to further investigate the potential of domain mapping for WT fault detection. In our study, hyperparameter tuning was evaluated on the test set of one selected domain pair using 1 month of target training data. It is possible that certain types of WT pairs exhibiting varying domain shifts may require differently weighted loss constraints and models, for which more generalizable optimization techniques could be investigated. Moreover, performance differences between WT pairs indicate a potential benefit in appropriately selecting suitable source-target WTs, which suggests needed future investigations to find appropriate selection criteria, e.g., based on the of the domain shift. Our domain mapping network furthermore exhibits a significantly higher complexity and a computationally heavier and longer training procedure compared to fine-tuning, which should be considered for practical applications. While we validated our approach on a very comprehensive dataset comprising real operational data from numerous WTs from different wind farms, our experiments were limited to WTs of the same manufacturer and a selection of SCADA features. Further large-scale experiments with more versatile wind farms and WT types, as well as different SCADA variables and systems, will be valuable to assess the generalizability of our results to other manufacturers. Lastly, evaluating our method when logs are available of various types of faults could provide insights into the limitations of mapping specific types of faults and lead to further improvements. 

Our exploratory work suggests a promising potential for further domain mapping applications. For WT fault detection, this includes investigating the use of different architectures and adjusted techniques to establish domain mapping as a novel and effective alternative to fine-tuning. The proposed technique moreover highlights a potential for applications beyond WT systems. Exploring its applicability to unsupervised anomaly detection-based tasks under data scarcity of other areas, such as for photovoltaic systems, is an interesting area of future research to explore domain mapping for more accurate and reliable models when faced with limited training data.

\section*{Acknowledgements}
This research was funded by the Swiss National Science Foundation (grant number 206342). We want to thank aventron AG (Weidenstrasse 27, 4142 Münchenstein, Switzerland, \url{www.aventron.com}) for sharing their measurement data enabling this research.

\printbibliography

% - - - -- - - -- - - -- APPENDIX 
% - - - -- - - -- - - -- APPENDIX 
% - - - -- - - -- - - -- APPENDIX 
% - - - -- - - -- - - -- APPENDIX 
% - - - -- - - -- - - -- APPENDIX 
% - - - -- - - -- - - -- APPENDIX 
% - - - -- - - -- - - -- APPENDIX 

\setcounter{table}{0}
\renewcommand{\thetable}{A\arabic{table}}

\setcounter{figure}{0}
\renewcommand{\thefigure}{A\arabic{figure}}

\newpage
\section*{Appendix A: Autoencoder-based NBM}
\label{sec:AppendixNBM}
All our autoencoder-based NBMs follow the model specifications outlined in Table A1, based on a model architecture search on a randomly chosen WT. Each NBM was trained using the RAdam \cite{RAdam} optimizer with a learning rate of 0.003, minimizing the mean squared reconstruction error between model input and autoencoder reconstruction. The models were trained using a batch size of 128 SCADA samples. Training was stopped once the reconstruction error on the validation set stopped improved for several data scarcity-dependent epochs (250 for 2 weeks, 25 for full target data). 

\input{Tables/ae_table}

\newpage
\section*{Appendix B: Domain Mapping Network}
\label{sec:AppendixDomainMapping}
\begin{figure}[h!]
    \centering
    \includegraphics[width=.3\textwidth]{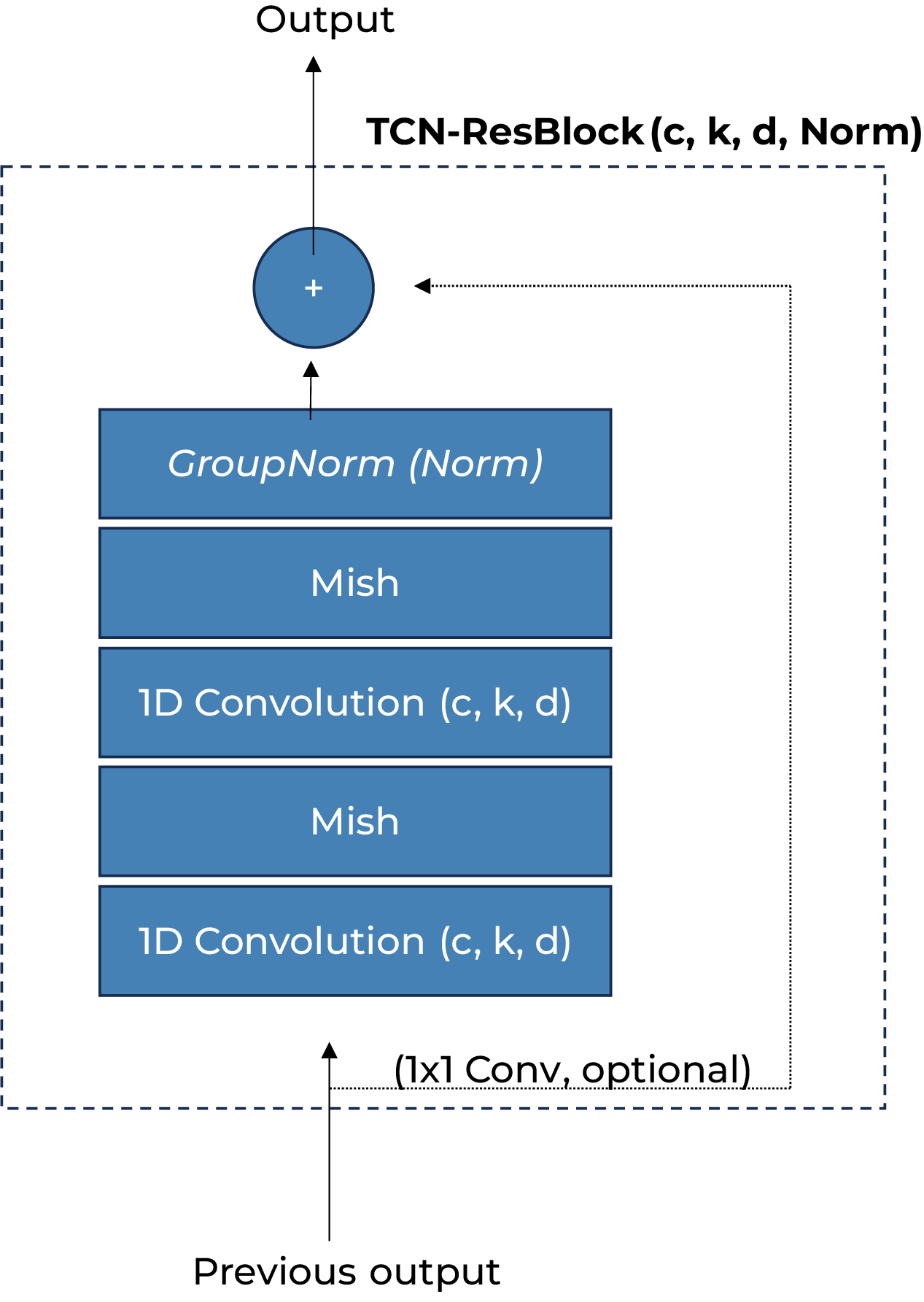}
    \caption{Structure of our TCN residual block, performing non-causal 1D convolutions with kernel size $k$, dilation $d$, and with $c$ channels. The final group normalization is optional when \textit{Norm} is set. Based on \cite{bai_empirical_2018}.}
    \label{fig:TCNBlock}
\end{figure}

The used architecture for the domain mapping discriminators is outlined in Table \ref{tab:DISC_TABLE}. For the generators, we used a residual temporal convolutional network (TCN) approach \cite{bai_empirical_2018}. Our non-causal architecture consists of several residual TCN blocks, kernel size, dilation, and normalization, depicted in Figure \ref{fig:TCNBlock}. The full generator is outlined in Table \ref{tab:TCN_Table}.

\input{Tables/DISC_TABLE}

\input{Tables/GAN_table}

\paragraph{GAN-QP.} We follow the GAN-QP formulation to train our generators and discriminators. Let $G$ be a generator, $T$ a discriminator, $x_r$ a real sample and $x_f$ a fake sample (in our case, a mapped sample). The GAN-QP loss is defined as \cite{suGANQPNovelGAN2018}:

%% taken & adjusted from the GAN-QP paper
\begin{equation}\begin{aligned}&T= \mathop{\arg\max}_T\, \mathbb{E}_{(x_r,x_f)\sim p(x_r)q(x_f)}\left[T(x_r)-T(x_f) - \frac{(T(x_r)-T(x_f))^2}{2\lambda_{QP} d(x_r,x_f)}\right] \\
&G = \mathop{\arg\min}_G\,\mathbb{E}_{(x_r,x_f)\sim p(x_r)q(x_f)}\left[T(x_r)-T(x_f)\right]
\end{aligned}\end{equation}

where in our study we use the Euclidean distance as $d$ and set $\lambda_{QP} = 1$. 

\paragraph{Training.} Our domain mapping network was trained with a batch size of 128 using an Adam optimizer ($\beta_1 = 0.5, \beta_2 = 0.999$, learning rate = $0.0002$) for the generators and discriminators each. Early stopping was implemented to stop training at an optimal state and to prevent overfitting. We use a reconstruction error obtained from the source NBM, by mapping the target validation data to the source domain. As only normal data is used for training and validation, we expect the reconstruction error of mapped target-to-source data to be low, representing realistic source WT data familiar to the source NBM. A rising reconstruction error could indicate overfitting on training data and unrealistic sample generation, as the mapped validation data starts to resemble the source domain less. It should be noted that this value cannot be used for model selection and tuning, as it contains no information regarding the content preservation (e.g., mapping all target samples to the same normal source domain sample would result in a very low score). Training was stopped after 1000 batch iterations with no improvement of the early stopping score. Moreover, at evaluation time, we used an Exponential Moving Average (EMA) of the training weights for the generators, which improved training stability and performance (as investigated in \cite{emagan}).  

\paragraph{Anomaly Augmentation.} The domain mapping network is trained to translate normal SCADA data resulting in the model inherently learning only the mapping relationships of normal data. During test time, when anomalous samples can be introduced, it may therefore lead to unwanted behavior in the sense that the mapping might "repair" anomalies or cause inconsistent mappings. We therefore add artificially anomalous data to the training set by duplicating each sample from the training batches with a random modification, namely by setting a random part of the 12 hours (40 - 100\%) and channels of a random feature group (e.g., power) to zero. Further investigations are needed to determine more augmentation techniques. While we found this step to be critical for performance, adding a further random scaling augmentation did not improve results.

\paragraph{Hyperparameters.} A source-target WT domain pair was randomly selected to search for a model architecture and optimal hyperparameters. The target WT used for optimization (WT07) was subsequently excluded as possible target domain from the evaluation. We considered multiple candidates using a target domain scarcity of 1 month and evaluated the F1-score of the resulting threshold scores with ones from an NBM trained on the full representative target domain training data. The same architecture and hyperparameters were used for all other domain pairs, although it is questionable whether these remain optimal across different data scarcity degrees and domain shifts, i.e., different deviations between the source and target WT. Setting hyperparameters based on the domain distance is subject to future research. 

The resulting full training procedure is described in Algorithm \ref{alg:1}. For more detailed specifications we refer to our provided implementation.

\paragraph{Implementation.} This work was implemented in PyTorch and trained using an NVIDIA GPU. Our code implementation is publicly available on GitHub \url{https://github.com/EnergyWeatherAI/WT_Generative_Domain_Adaptation}.

\input{Algorithms/algo1}

\newpage
\section*{Appendix C: Detailed Results}
\input{Tables/f1tablefull}

\newpage 
\section*{Appendix D: WT Overview}

\input{Tables/datatable}

\label{sec:AppendixWTComparison}
\begin{figure}[h!]
    \centering
    \includegraphics[width=\textwidth]{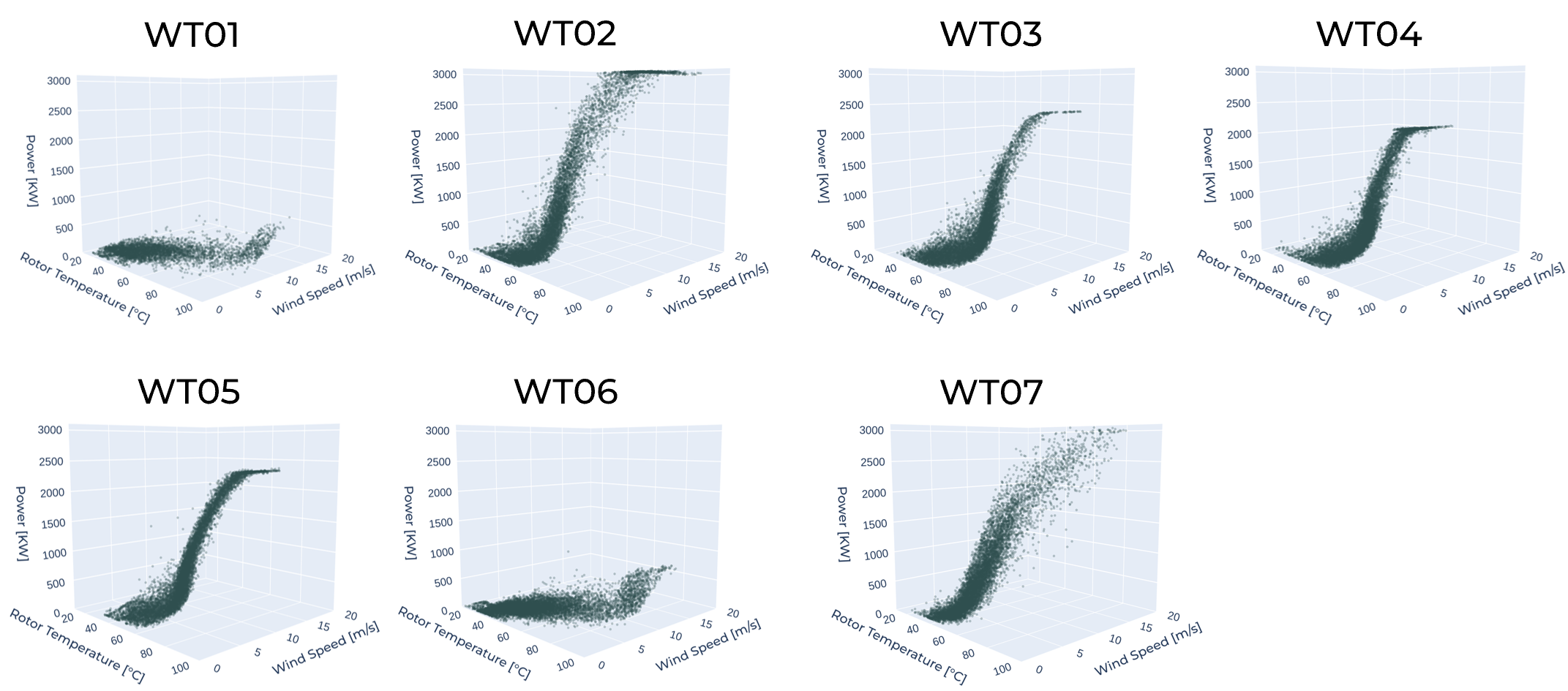}
    \caption{Scatter plots of the mean power, wind speed, and rotor temperature of a subset from filtered training samples illustrating the differences across all 7 WTs used in our experiments.}
    \label{fig:wt_overview}
\end{figure}

\newpage
\section*{Appendix E: Representative Training Data}
\label{sec:AppendixRepresentative}
\begin{figure}[h!]
    \centering
    \includegraphics[width=.8\textwidth]{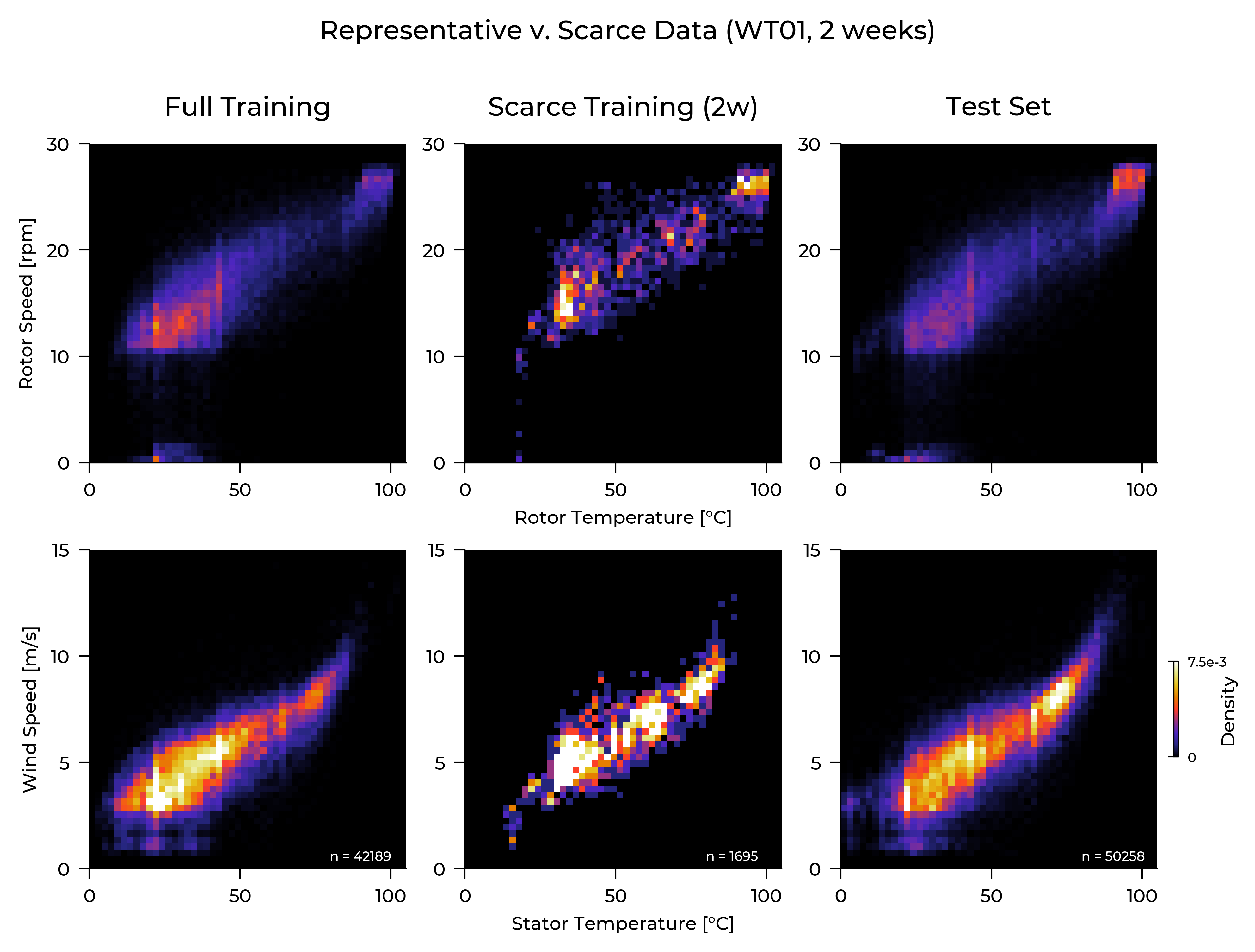}
    \caption{Density 2D-histograms with SCADA data of WT01. The top row shows the relationship between the rotor temperature and the rotor speed for the full training data (left column), for a scarcity scenario of 2 weeks of training data (middle column), and the filtered test set without incidents (right column). The bottom row visualizes the respective relationships between the wind speed and the stator temperature. We observe a visually very high similarity between the histograms of the full training data and the test set, i.e., the full training data appears to be \textit{representative} of the WT's operational states. On the other hand, the histograms based on scarce training data show noticeable deviations to the test set data. For instance, a narrower range and less variability for low temperatures and low rotor and wind speeds.}
    \label{fig:representative_data}
\end{figure}

\end{document}

%% file: Tables/mean_f1_scores.tex
% Please add the following required packages to your document preamble:
% \usepackage{booktabs}
\begin{table}[ht!]
\centering
\caption{F1-scores (in \%) across all 6 target wind turbines evaluating the threshold similarity of NBMs trained on various degrees of limited training data to the ground truth of the respective NBMs trained on a full training dataset.}
\label{tab:mean_f1_scores}
\begin{tabular}{@{}lllllll@{}}
\toprule
\textbf{Target WT} &
  \multicolumn{1}{c}{\textbf{2 months}} &
  \multicolumn{1}{c}{\textbf{6 weeks}} &
  \multicolumn{1}{c}{\textbf{1 month}} &
  \multicolumn{1}{c}{\textbf{3 weeks}} &
  \multicolumn{1}{c}{\textbf{2 weeks}} &
  \multicolumn{1}{c}{\textbf{1 week}} \\ \midrule
WT01 & 89.0 & 84.6 & 77.4 & 74.4 & 77.1 & 66.4 \\
WT02 & 77.8 & 62.6 & 65.3 & 60.3 & 73.6 & 35.2 \\
WT03 & 84.7 & 73.8 & 64.8 & 38.4 & 52.6 & 34.5 \\
WT04 & 92.6 & 94.0 & 88.3 & 89.8 & 42.8 & 34.4 \\
WT05 & 67.6 & 55.3 & 63.9 & 56.5 & 31.5 & 24.3 \\
WT06 & 94.5 & 42.4 & 24.7 & 31.1 & 30.3 & 7.0  \\ \bottomrule
\end{tabular}
\end{table}

%% file: Tables/f1_score_diffs.tex
% Please add the following required packages to your document preamble:
% \usepackage{booktabs}
% \usepackage{multirow}
% \usepackage{graphicx}
% \usepackage[table,xcdraw]{xcolor}
% Beamer presentation requires \usepackage{colortbl} instead of \usepackage[table,xcdraw]{xcolor}
\begin{table}[ht!]
\centering
\caption{Change in F1-score value compared to the respective NBM trained on scarce data only for all 28 domain pairs and data scarcity degrees.}
\label{tab:f1_score_diffs}
\resizebox{\textwidth}{!}{%
\begin{tabular}{@{}cl|ll|ll|ll|ll|ll|ll@{}}
\toprule
\rowcolor[HTML]{FFFFFF} 
\multicolumn{1}{l}{\cellcolor[HTML]{FFFFFF}} &
   &
  \multicolumn{2}{c|}{\cellcolor[HTML]{FFFFFF}2 months} &
  \multicolumn{2}{c|}{\cellcolor[HTML]{FFFFFF}6 weeks} &
  \multicolumn{2}{c|}{\cellcolor[HTML]{FFFFFF}1 month} &
  \multicolumn{2}{c|}{\cellcolor[HTML]{FFFFFF}3 weeks} &
  \multicolumn{2}{c|}{\cellcolor[HTML]{FFFFFF}2 weeks} &
  \multicolumn{2}{c}{\cellcolor[HTML]{FFFFFF}1 week} \\ \midrule
\textbf{Target} &
  \multicolumn{1}{c|}{\textbf{Source}} &
  \multicolumn{1}{c}{\textbf{FT}} &
  \multicolumn{1}{c|}{\textbf{Ours}} &
  \multicolumn{1}{c}{\textbf{FT}} &
  \multicolumn{1}{c|}{\textbf{Ours}} &
  \multicolumn{1}{c}{\textbf{FT}} &
  \multicolumn{1}{c|}{\textbf{Ours}} &
  \multicolumn{1}{c}{\textbf{FT}} &
  \multicolumn{1}{c|}{\textbf{Ours}} &
  \multicolumn{1}{c}{\textbf{FT}} &
  \multicolumn{1}{c|}{\textbf{Ours}} &
  \multicolumn{1}{c}{\textbf{FT}} &
  \multicolumn{1}{c}{\textbf{Ours}} \\ \midrule
 &
  WT07 &
  \cellcolor[HTML]{FFFFFF}+0.1 &
  \cellcolor[HTML]{FEFBFB}-1.1 &
  \cellcolor[HTML]{F6FCF8}+3.9 &
  \cellcolor[HTML]{FAFDFB}+2.1 &
  \cellcolor[HTML]{FFFFFF}+0.0 &
  \cellcolor[HTML]{EBF7EE}+8.6 &
  \cellcolor[HTML]{FEFBFB}-1.3 &
  \cellcolor[HTML]{FCCACA}-17.1 &
  \cellcolor[HTML]{FEF5F5}-3.1 &
  \cellcolor[HTML]{FAFDFB}+2.3 &
  \cellcolor[HTML]{EBF7EF}+8.3 &
  \cellcolor[HTML]{E2F3E6}+12.4 \\
 &
  WT02 &
  \cellcolor[HTML]{FA9B9C}-32.2 &
  \cellcolor[HTML]{FBB3B4}-24.6 &
  \cellcolor[HTML]{FDE8E9}-7.2 &
  \cellcolor[HTML]{FBFEFC}+1.7 &
  \cellcolor[HTML]{FEEEEE}-5.3 &
  \cellcolor[HTML]{E2F3E7}+12.1 &
  \cellcolor[HTML]{FEF2F2}-4.1 &
  \cellcolor[HTML]{F3FAF5}+5.2 &
  \cellcolor[HTML]{FEF1F2}-4.2 &
  \cellcolor[HTML]{F1F9F3}+6.0 &
  \cellcolor[HTML]{F1FAF3}+6.0 &
  \cellcolor[HTML]{E7F5EB}+10.3 \\
 &
  WT05 &
  \cellcolor[HTML]{FEEDED}-5.8 &
  \cellcolor[HTML]{FEEAEA}-6.7 &
  \cellcolor[HTML]{FEF3F3}-3.7 &
  \cellcolor[HTML]{F7FCF8}+3.7 &
  \cellcolor[HTML]{FEF3F3}-3.8 &
  \cellcolor[HTML]{F0F9F2}+6.6 &
  \cellcolor[HTML]{FEFFFE}+0.7 &
  \cellcolor[HTML]{DFF2E4}+13.6 &
  \cellcolor[HTML]{FBFEFC}+1.8 &
  \cellcolor[HTML]{F0F9F2}+6.4 &
  \cellcolor[HTML]{EAF7EE}+8.9 &
  \cellcolor[HTML]{E9F6ED}+9.2 \\
\multirow{-4}{*}{\textbf{WT01}} &
  WT04 &
  \cellcolor[HTML]{FEF3F3}-3.8 &
  \cellcolor[HTML]{FBFEFC}+1.7 &
  \cellcolor[HTML]{FEF2F2}-3.9 &
  \cellcolor[HTML]{F2FAF4}+5.8 &
  \cellcolor[HTML]{FDFFFE}+0.9 &
  \cellcolor[HTML]{F0F9F2}+6.6 &
  \cellcolor[HTML]{FEF6F6}-2.9 &
  \cellcolor[HTML]{D8EFDE}+16.4 &
  \cellcolor[HTML]{FEF1F1}-4.4 &
  \cellcolor[HTML]{E3F4E8}+11.8 &
  \cellcolor[HTML]{FFFFFF}+0.3 &
  \cellcolor[HTML]{E2F3E7}+12.1 \\ \midrule
 &
  WT01 &
  \cellcolor[HTML]{F9FDFA}+2.8 &
  \cellcolor[HTML]{FBA9AA}-27.7 &
  \cellcolor[HTML]{CDEAD5}+21.1 &
  \cellcolor[HTML]{FBFEFC}+2.0 &
  \cellcolor[HTML]{D2EDD9}+19.0 &
  \cellcolor[HTML]{FDE3E3}-9.0 &
  \cellcolor[HTML]{DFF2E4}+13.6 &
  \cellcolor[HTML]{FEF3F3}-3.7 &
  \cellcolor[HTML]{F98082}-40.9 &
  \cellcolor[HTML]{FA9FA0}-31.1 &
  \cellcolor[HTML]{FA9899}-33.3 &
  \cellcolor[HTML]{EFF9F2}+6.7 \\
 &
  WT06 &
  \cellcolor[HTML]{FCCECF}-15.6 &
  \cellcolor[HTML]{FEECEC}-6.1 &
  \cellcolor[HTML]{D4EDDA}+18.3 &
  \cellcolor[HTML]{F0F9F3}+6.4 &
  \cellcolor[HTML]{D3EDDA}+18.6 &
  \cellcolor[HTML]{DCF1E2}+14.6 &
  \cellcolor[HTML]{DCF1E2}+14.6 &
  \cellcolor[HTML]{D7EFDD}+16.8 &
  \cellcolor[HTML]{FA999A}-32.9 &
  \cellcolor[HTML]{FDE3E3}-8.9 &
  \cellcolor[HTML]{FBAEAF}-26.1 &
  \cellcolor[HTML]{BFE5C9}+26.9 \\
 &
  WT05 &
  \cellcolor[HTML]{F5FBF7}+4.4 &
  \cellcolor[HTML]{FCD2D3}-14.3 &
  \cellcolor[HTML]{D4EDDA}+18.2 &
  \cellcolor[HTML]{FBFEFC}+1.8 &
  \cellcolor[HTML]{E9F6ED}+9.2 &
  \cellcolor[HTML]{FEFEFE}-0.2 &
  \cellcolor[HTML]{D6EEDC}+17.4 &
  \cellcolor[HTML]{F3FAF5}+5.2 &
  \cellcolor[HTML]{EAF7ED}+9.0 &
  \cellcolor[HTML]{FDD8D9}-12.4 &
  \cellcolor[HTML]{A9DBB6}+36.1 &
  \cellcolor[HTML]{C0E5CA}+26.3 \\
\multirow{-4}{*}{\textbf{WT02}} &
  WT04 &
  \cellcolor[HTML]{EFF9F1}+7.0 &
  \cellcolor[HTML]{FDD6D7}-13.0 &
  \cellcolor[HTML]{CCEAD4}+21.3 &
  \cellcolor[HTML]{F5FBF7}+4.4 &
  \cellcolor[HTML]{EEF8F1}+7.3 &
  \cellcolor[HTML]{FBFEFC}+1.8 &
  \cellcolor[HTML]{D3EDDA}+18.3 &
  \cellcolor[HTML]{E7F5EA}+10.4 &
  \cellcolor[HTML]{EFF8F1}+7.0 &
  \cellcolor[HTML]{FDDFDF}-10.2 &
  \cellcolor[HTML]{B5E1C1}+30.9 &
  \cellcolor[HTML]{D0ECD7}+19.9 \\ \midrule
 &
  WT01 &
  \cellcolor[HTML]{F8696B}-48.6 &
  \cellcolor[HTML]{FBFEFC}+1.8 &
  \cellcolor[HTML]{F9FDFA}+2.8 &
  \cellcolor[HTML]{E0F3E5}+13.0 &
  \cellcolor[HTML]{FDE6E6}-7.9 &
  \cellcolor[HTML]{D0ECD7}+19.9 &
  \cellcolor[HTML]{A3D9B2}+38.4 &
  \cellcolor[HTML]{ACDDB9}+34.9 &
  \cellcolor[HTML]{DDF1E2}+14.5 &
  \cellcolor[HTML]{CDEBD5}+20.9 &
  \cellcolor[HTML]{FCC9C9}-17.4 &
  \cellcolor[HTML]{F1F9F3}+6.2 \\
 &
  WT06 &
  \cellcolor[HTML]{F97F80}-41.5 &
  \cellcolor[HTML]{FEEDEE}-5.5 &
  \cellcolor[HTML]{F4FBF6}+4.6 &
  \cellcolor[HTML]{F2FAF4}+5.8 &
  \cellcolor[HTML]{FAFDFB}+2.2 &
  \cellcolor[HTML]{EEF8F1}+7.2 &
  \cellcolor[HTML]{A7DBB5}+36.8 &
  \cellcolor[HTML]{FBFDFB}+2.1 &
  \cellcolor[HTML]{E7F5EB}+10.1 &
  \cellcolor[HTML]{B1DFBD}+32.7 &
  \cellcolor[HTML]{FCCBCC}-16.7 &
  \cellcolor[HTML]{EEF8F1}+7.2 \\
 &
  WT07 &
  \cellcolor[HTML]{FA9E9F}-31.4 &
  \cellcolor[HTML]{FAFDFB}+2.1 &
  \cellcolor[HTML]{FEF8F8}-2.1 &
  \cellcolor[HTML]{E3F4E7}+11.8 &
  \cellcolor[HTML]{DDF1E2}+14.4 &
  \cellcolor[HTML]{FCCDCD}-16.2 &
  \cellcolor[HTML]{92D2A3}+45.7 &
  \cellcolor[HTML]{EDF8F0}+7.6 &
  \cellcolor[HTML]{C7E8CF}+23.6 &
  \cellcolor[HTML]{A7DBB5}+36.6 &
  \cellcolor[HTML]{CFEBD6}+20.3 &
  \cellcolor[HTML]{DAF0E0}+15.7 \\
 &
  WT02 &
  \cellcolor[HTML]{FA9E9F}-31.4 &
  \cellcolor[HTML]{FEEBEC}-6.2 &
  \cellcolor[HTML]{F3FAF5}+5.1 &
  \cellcolor[HTML]{E8F6EC}+9.6 &
  \cellcolor[HTML]{CCEAD4}+21.6 &
  \cellcolor[HTML]{FCCDCD}-16.1 &
  \cellcolor[HTML]{ABDCB8}+35.3 &
  \cellcolor[HTML]{A5DAB3}+37.8 &
  \cellcolor[HTML]{DEF2E3}+13.8 &
  \cellcolor[HTML]{C1E5CB}+26.0 &
  \cellcolor[HTML]{F0F9F2}+6.5 &
  \cellcolor[HTML]{D4EEDB}+17.9 \\
 &
  WT05 &
  \cellcolor[HTML]{F5FBF7}+4.2 &
  \cellcolor[HTML]{FDD9D9}-12.2 &
  \cellcolor[HTML]{D9EFDF}+16.1 &
  \cellcolor[HTML]{E3F4E8}+11.7 &
  \cellcolor[HTML]{E3F3E7}+12.0 &
  \cellcolor[HTML]{D5EEDC}+17.6 &
  \cellcolor[HTML]{9ED7AD}+40.5 &
  \cellcolor[HTML]{8ED09F}+47.2 &
  \cellcolor[HTML]{A9DBB6}+36.1 &
  \cellcolor[HTML]{CEEBD6}+20.6 &
  \cellcolor[HTML]{78C78D}+56.3 &
  \cellcolor[HTML]{BFE4C9}+27.0 \\
\multirow{-6}{*}{\textbf{WT03}} &
  WT04 &
  \cellcolor[HTML]{FEF5F5}-3.0 &
  \cellcolor[HTML]{FEE9E9}-6.9 &
  \cellcolor[HTML]{D9F0DF}+16.0 &
  \cellcolor[HTML]{F2FAF4}+5.5 &
  \cellcolor[HTML]{E8F6EB}+9.9 &
  \cellcolor[HTML]{FDFEFD}+1.0 &
  \cellcolor[HTML]{98D4A8}+43.3 &
  \cellcolor[HTML]{90D1A2}+46.2 &
  \cellcolor[HTML]{B0DEBC}+33.2 &
  \cellcolor[HTML]{D6EEDC}+17.3 &
  \cellcolor[HTML]{B6E1C1}+30.6 &
  \cellcolor[HTML]{DDF1E3}+14.2 \\ \midrule
 &
  WT01 &
  \cellcolor[HTML]{FDD7D8}-12.7 &
  \cellcolor[HTML]{FCCACA}-17.2 &
  \cellcolor[HTML]{FBB7B8}-23.1 &
  \cellcolor[HTML]{FCD0D0}-15.1 &
  \cellcolor[HTML]{FDE1E2}-9.5 &
  \cellcolor[HTML]{FEF4F5}-3.3 &
  \cellcolor[HTML]{FBB9BA}-22.5 &
  \cellcolor[HTML]{FDE7E7}-7.6 &
  \cellcolor[HTML]{FEFDFD}-0.5 &
  \cellcolor[HTML]{E3F4E8}+11.7 &
  \cellcolor[HTML]{FAFDFB}+2.4 &
  \cellcolor[HTML]{EAF7ED}+9.0 \\
 &
  WT06 &
  \cellcolor[HTML]{FEEBEC}-6.2 &
  \cellcolor[HTML]{FDDFDF}-10.2 &
  \cellcolor[HTML]{FDE5E5}-8.3 &
  \cellcolor[HTML]{FBBCBD}-21.7 &
  \cellcolor[HTML]{FDE8E8}-7.3 &
  \cellcolor[HTML]{FDE6E7}-7.8 &
  \cellcolor[HTML]{FBB0B1}-25.6 &
  \cellcolor[HTML]{FEF5F5}-3.1 &
  \cellcolor[HTML]{F9FDFA}+2.5 &
  \cellcolor[HTML]{EFF9F1}+6.9 &
  \cellcolor[HTML]{ECF8EF}+7.9 &
  \cellcolor[HTML]{E1F3E6}+12.6 \\
 &
  WT07 &
  \cellcolor[HTML]{FEF2F2}-4.0 &
  \cellcolor[HTML]{FDE5E5}-8.3 &
  \cellcolor[HTML]{FEF9F9}-1.7 &
  \cellcolor[HTML]{FDE2E3}-9.1 &
  \cellcolor[HTML]{FCFEFC}+1.6 &
  \cellcolor[HTML]{FEEDED}-5.8 &
  \cellcolor[HTML]{FEEAEB}-6.5 &
  \cellcolor[HTML]{FDDDDD}-10.9 &
  \cellcolor[HTML]{D5EEDC}+17.6 &
  \cellcolor[HTML]{C7E8D0}+23.5 &
  \cellcolor[HTML]{C8E9D1}+22.9 &
  \cellcolor[HTML]{BFE5C9}+26.8 \\
 &
  WT02 &
  \cellcolor[HTML]{FEF2F2}-4.1 &
  \cellcolor[HTML]{FEFEFE}-0.3 &
  \cellcolor[HTML]{FDD7D7}-12.9 &
  \cellcolor[HTML]{FEF1F1}-4.4 &
  \cellcolor[HTML]{FEFDFD}-0.5 &
  \cellcolor[HTML]{F8FCF9}+3.2 &
  \cellcolor[HTML]{FDE9E9}-7.1 &
  \cellcolor[HTML]{FEF1F1}-4.5 &
  \cellcolor[HTML]{CEEBD6}+20.5 &
  \cellcolor[HTML]{BEE4C8}+27.1 &
  \cellcolor[HTML]{D4EDDB}+18.0 &
  \cellcolor[HTML]{BEE4C8}+27.1 \\
\multirow{-5}{*}{\textbf{WT04}} &
  WT05 &
  \cellcolor[HTML]{FEFCFC}-0.9 &
  \cellcolor[HTML]{FDFFFE}+0.9 &
  \cellcolor[HTML]{FEF8F8}-2.2 &
  \cellcolor[HTML]{FEFDFD}-0.4 &
  \cellcolor[HTML]{F5FBF6}+4.5 &
  \cellcolor[HTML]{F0F9F2}+6.4 &
  \cellcolor[HTML]{FAFDFB}+2.3 &
  \cellcolor[HTML]{F7FCF8}+3.7 &
  \cellcolor[HTML]{95D3A5}+44.5 &
  \cellcolor[HTML]{D4EDDA}+18.2 &
  \cellcolor[HTML]{71C487}+59.2 &
  \cellcolor[HTML]{BAE3C5}+29.0 \\ \midrule
 &
  WT01 &
  \cellcolor[HTML]{FEF7F7}-2.5 &
  \cellcolor[HTML]{FEEDED}-5.7 &
  \cellcolor[HTML]{E8F6EC}+9.6 &
  \cellcolor[HTML]{FEF3F3}-3.7 &
  \cellcolor[HTML]{FDDFE0}-10.1 &
  \cellcolor[HTML]{FEFAFA}-1.5 &
  \cellcolor[HTML]{FDDFE0}-10.1 &
  \cellcolor[HTML]{FDDCDC}-11.3 &
  \cellcolor[HTML]{FEEDEE}-5.5 &
  \cellcolor[HTML]{FBFEFC}+1.7 &
  \cellcolor[HTML]{EDF8F0}+7.6 &
  \cellcolor[HTML]{FEF0F0}-4.8 \\
 &
  WT06 &
  \cellcolor[HTML]{FEF9F9}-1.8 &
  \cellcolor[HTML]{E6F5EA}+10.7 &
  \cellcolor[HTML]{ECF7EF}+8.2 &
  \cellcolor[HTML]{CCEAD4}+21.3 &
  \cellcolor[HTML]{FDE3E3}-9.1 &
  \cellcolor[HTML]{EFF9F2}+6.8 &
  \cellcolor[HTML]{FEEBEB}-6.5 &
  \cellcolor[HTML]{C5E7CE}+24.3 &
  \cellcolor[HTML]{F9FDFA}+2.6 &
  \cellcolor[HTML]{8ED0A0}+47.2 &
  \cellcolor[HTML]{FDDCDC}-11.2 &
  \cellcolor[HTML]{C8E8D1}+23.1 \\
 &
  WT07 &
  \cellcolor[HTML]{D0ECD7}+19.8 &
  \cellcolor[HTML]{ECF7EF}+8.1 &
  \cellcolor[HTML]{BCE3C6}+28.2 &
  \cellcolor[HTML]{CDEBD5}+20.8 &
  \cellcolor[HTML]{F4FBF6}+4.9 &
  \cellcolor[HTML]{E1F3E6}+12.6 &
  \cellcolor[HTML]{E1F3E6}+12.6 &
  \cellcolor[HTML]{CAE9D2}+22.3 &
  \cellcolor[HTML]{B0DFBD}+32.9 &
  \cellcolor[HTML]{D8EFDE}+16.4 &
  \cellcolor[HTML]{A8DBB5}+36.5 &
  \cellcolor[HTML]{BDE4C7}+27.8 \\
 &
  WT02 &
  \cellcolor[HTML]{D0ECD7}+19.8 &
  \cellcolor[HTML]{DDF1E3}+14.2 &
  \cellcolor[HTML]{B5E0C0}+31.2 &
  \cellcolor[HTML]{C2E6CC}+25.4 &
  \cellcolor[HTML]{D1ECD8}+19.3 &
  \cellcolor[HTML]{CAE9D2}+22.2 &
  \cellcolor[HTML]{CFEBD7}+20.1 &
  \cellcolor[HTML]{B2DFBE}+32.4 &
  \cellcolor[HTML]{9FD7AE}+40.2 &
  \cellcolor[HTML]{B4E0C0}+31.3 &
  \cellcolor[HTML]{C3E6CD}+25.1 &
  \cellcolor[HTML]{BAE3C5}+28.8 \\
\multirow{-5}{*}{\textbf{WT05}} &
  WT04 &
  \cellcolor[HTML]{F1F9F3}+6.1 &
  \cellcolor[HTML]{C2E6CB}+25.6 &
  \cellcolor[HTML]{E7F5EB}+10.2 &
  \cellcolor[HTML]{B1DFBD}+32.8 &
  \cellcolor[HTML]{EFF9F2}+6.9 &
  \cellcolor[HTML]{E9F6EC}+9.5 &
  \cellcolor[HTML]{E3F4E8}+11.8 &
  \cellcolor[HTML]{BFE4C9}+27.0 &
  \cellcolor[HTML]{85CC98}+51.0 &
  \cellcolor[HTML]{8CCF9E}+48.0 &
  \cellcolor[HTML]{BFE5C9}+26.9 &
  \cellcolor[HTML]{B0DEBC}+33.3 \\ \midrule
 &
  WT07 &
  \cellcolor[HTML]{FBABAC}-27.2 &
  \cellcolor[HTML]{FAA7A8}-28.3 &
  \cellcolor[HTML]{D7EFDD}+16.8 &
  \cellcolor[HTML]{D3EDDA}+18.4 &
  \cellcolor[HTML]{D9EFDF}+16.1 &
  \cellcolor[HTML]{63BE7B}+64.9 &
  \cellcolor[HTML]{F3FAF5}+5.2 &
  \cellcolor[HTML]{D3EDDA}+18.6 &
  \cellcolor[HTML]{EDF8F0}+7.7 &
  \cellcolor[HTML]{CEEBD5}+20.8 &
  \cellcolor[HTML]{FEFCFC}-0.7 &
  \cellcolor[HTML]{F6FCF8}+3.8 \\
 &
  WT02 &
  \cellcolor[HTML]{FAA0A1}-30.8 &
  \cellcolor[HTML]{FCC7C8}-17.8 &
  \cellcolor[HTML]{D4EDDB}+18.2 &
  \cellcolor[HTML]{B4E0C0}+31.3 &
  \cellcolor[HTML]{D5EEDB}+17.8 &
  \cellcolor[HTML]{88CE9A}+49.7 &
  \cellcolor[HTML]{ECF7EF}+8.2 &
  \cellcolor[HTML]{D9F0DF}+16.0 &
  \cellcolor[HTML]{FAFDFB}+2.3 &
  \cellcolor[HTML]{E8F6EC}+9.8 &
  \cellcolor[HTML]{FEFCFC}-1.0 &
  \cellcolor[HTML]{FEFEFE}-0.1 \\
 &
  WT05 &
  \cellcolor[HTML]{FCC2C3}-19.6 &
  \cellcolor[HTML]{FFFFFF}+0.4 &
  \cellcolor[HTML]{FEFFFE}+0.5 &
  \cellcolor[HTML]{DAF0E0}+15.4 &
  \cellcolor[HTML]{FEF2F2}-4.0 &
  \cellcolor[HTML]{B9E2C4}+29.3 &
  \cellcolor[HTML]{FDD7D8}-12.8 &
  \cellcolor[HTML]{94D3A5}+44.6 &
  \cellcolor[HTML]{FDDDDE}-10.8 &
  \cellcolor[HTML]{A0D8AF}+39.7 &
  \cellcolor[HTML]{FEFCFC}-0.9 &
  \cellcolor[HTML]{FEFAFA}-1.3 \\
\multirow{-4}{*}{\textbf{WT06}} &
  WT04 &
  \cellcolor[HTML]{FCD0D0}-15.2 &
  \cellcolor[HTML]{FFFFFF}+0.4 &
  \cellcolor[HTML]{D6EEDC}+17.3 &
  \cellcolor[HTML]{AEDDBA}+34.1 &
  \cellcolor[HTML]{F7FCF9}+3.4 &
  \cellcolor[HTML]{8DD09F}+47.5 &
  \cellcolor[HTML]{FCD0D0}-15.1 &
  \cellcolor[HTML]{94D3A5}+44.8 &
  \cellcolor[HTML]{FDE4E4}-8.7 &
  \cellcolor[HTML]{8BCF9D}+48.6 &
  \cellcolor[HTML]{FEFBFB}-1.1 &
  \cellcolor[HTML]{FFFFFF}+0.0 \\ \midrule
\rowcolor[HTML]{FFFFFF} 
\multicolumn{2}{c|}{\cellcolor[HTML]{FFFFFF}} &
  \multicolumn{1}{c}{\cellcolor[HTML]{FFFFFF}-9.8} &
  \multicolumn{1}{c|}{\cellcolor[HTML]{FFFFFF}\textbf{-5.2}} &
  \multicolumn{1}{c}{\cellcolor[HTML]{FFFFFF}+7.2} &
  \multicolumn{1}{c|}{\cellcolor[HTML]{FFFFFF}\textbf{+8.2}} &
  \multicolumn{1}{c}{\cellcolor[HTML]{FFFFFF}+4.7} &
  \multicolumn{1}{c|}{\cellcolor[HTML]{FFFFFF}\textbf{+10.3}} &
  \multicolumn{1}{c}{\cellcolor[HTML]{FFFFFF}+8.9} &
  \multicolumn{1}{c|}{\cellcolor[HTML]{FFFFFF}\textbf{+15.0}} &
  \multicolumn{1}{c}{\cellcolor[HTML]{FFFFFF}+9.3} &
  \multicolumn{1}{c|}{\cellcolor[HTML]{FFFFFF}\textbf{+16.8}} &
  \multicolumn{1}{c}{\cellcolor[HTML]{FFFFFF}+10.8} &
  \multicolumn{1}{c}{\cellcolor[HTML]{FFFFFF}\textbf{+15.2}} \\
\rowcolor[HTML]{FFFFFF} 
\multicolumn{2}{c|}{\multirow{-2}{*}{\cellcolor[HTML]{FFFFFF}\textbf{\begin{tabular}[c]{@{}c@{}}Average\\ {[}+- std.{]}\end{tabular}}}} &
  \multicolumn{1}{c}{\cellcolor[HTML]{FFFFFF}{[}17.4{]}} &
  \multicolumn{1}{c|}{\cellcolor[HTML]{FFFFFF}{[}12.2{]}} &
  \multicolumn{1}{c}{\cellcolor[HTML]{FFFFFF}{[}12.9{]}} &
  \multicolumn{1}{c|}{\cellcolor[HTML]{FFFFFF}{[}13.6{]}} &
  \multicolumn{1}{c}{\cellcolor[HTML]{FFFFFF}{[}9.9{]}} &
  \multicolumn{1}{c|}{\cellcolor[HTML]{FFFFFF}{[}18.9{]}} &
  \multicolumn{1}{c}{\cellcolor[HTML]{FFFFFF}{[}20.3{]}} &
  \multicolumn{1}{c|}{\cellcolor[HTML]{FFFFFF}{[}19.1{]}} &
  \multicolumn{1}{c}{\cellcolor[HTML]{FFFFFF}{[}21.4{]}} &
  \multicolumn{1}{c|}{\cellcolor[HTML]{FFFFFF}{[}19.2{]}} &
  \multicolumn{1}{c}{\cellcolor[HTML]{FFFFFF}{[}22.3{]}} &
  \multicolumn{1}{c}{\cellcolor[HTML]{FFFFFF}{[}11.0{]}} \\ \bottomrule
\end{tabular}%
}
\end{table}

%% file: Tables/ae_table.tex
% Please add the following required packages to your document preamble:
% \usepackage{booktabs}
% \usepackage[table,xcdraw]{xcolor}
% Beamer presentation requires \usepackage{colortbl} instead of \usepackage[table,xcdraw]{xcolor}
\begin{table}[htb!]
\centering
\caption{Architecture of the autoencoder-based NBMs used in our study.}
\label{tab:ae_table}
\begin{tabular}{@{}
>{\columncolor[HTML]{FFFFFF}}l l@{}}
\multicolumn{2}{c}{\cellcolor[HTML]{FFFFFF}\textbf{Autoencoder model architecture}}                                                               \\ \midrule
\multicolumn{1}{l|}{\cellcolor[HTML]{FFFFFF}\textit{Input}}      & \cellcolor[HTML]{F2F2F2}\textit{11 channels x 72 datapoints}                   \\
\multicolumn{1}{l|}{\cellcolor[HTML]{FFFFFF}\textit{Block 1}}    & Conv1d   (32 filters, kernel size 7, stride 1, Mish \cite{misraMishSelfRegularized2019}) x 2              \\
\multicolumn{1}{l|}{\cellcolor[HTML]{FFFFFF}}                    & \cellcolor[HTML]{F2F2F2}MaxPool (kernel size 2)                                \\
\multicolumn{1}{l|}{\cellcolor[HTML]{FFFFFF}}                    & GroupNorm \cite{Wu_2018_ECCV} (1 group, 32 channels)                                   \\
\multicolumn{1}{l|}{\cellcolor[HTML]{FFFFFF}\textit{Block 2}}   & \cellcolor[HTML]{F2F2F2}Conv1d (32 filters, kernel size 5, stride 1, Mish) x 2 \\
\multicolumn{1}{l|}{\cellcolor[HTML]{FFFFFF}}                    & MaxPool   (kernel size 2)                                                      \\
\multicolumn{1}{l|}{\cellcolor[HTML]{FFFFFF}}                    & \cellcolor[HTML]{F2F2F2}GroupNorm  (1   group, 32 channels)                    \\
\multicolumn{1}{l|}{\cellcolor[HTML]{FFFFFF}\textit{Bottleneck}} & Flatten,   FC (32 x 18, 72), FC (72, 8 x 18)                             \\
\multicolumn{1}{l|}{\cellcolor[HTML]{FFFFFF}\textit{Block 3}}    & \cellcolor[HTML]{F2F2F2}Upsample (factor 2)             \\
\multicolumn{1}{l|}{\cellcolor[HTML]{FFFFFF}}                    & Conv1d   (32 filters, kernel size 3, stride 1, Mish) x 2                       \\
\multicolumn{1}{l|}{\cellcolor[HTML]{FFFFFF}}                    & \cellcolor[HTML]{F2F2F2}GroupNorm  (1   group, 32 channels)                    \\
\multicolumn{1}{l|}{\cellcolor[HTML]{FFFFFF}\textit{Block 4}}    & Upsample   (factor 2)                                   \\
\multicolumn{1}{l|}{\cellcolor[HTML]{FFFFFF}}                    & \cellcolor[HTML]{F2F2F2}Conv1d (32 filters, kernel size 3, stride 1, Mish) x 2 \\
\multicolumn{1}{l|}{\cellcolor[HTML]{FFFFFF}}                    & GroupNorm  (1 group, 32 channels)                                              \\
\multicolumn{1}{l|}{\cellcolor[HTML]{FFFFFF}\textit{Block Out}} & \cellcolor[HTML]{F2F2F2}Conv1d (11 filters, kernel size   1, stride 1, linear) \\
\multicolumn{1}{l|}{\cellcolor[HTML]{FFFFFF}\textit{Output}}     & \textit{11   channels x 72 datapoints}                                        
\end{tabular}
\end{table}

%% file: Tables/DISC_TABLE.tex
% Please add the following required packages to your document preamble:
% \usepackage[table,xcdraw]{xcolor}
% Beamer presentation requires \usepackage{colortbl} instead of \usepackage[table,xcdraw]{xcolor}
\begin{table}[ht!]
\centering
\caption{Description of our discriminator model architecture.}
\label{tab:DISC_TABLE}
\begin{tabular}{ll}
\multicolumn{2}{c}{\textbf{Discriminator   architecture}}                                            \\ \hline
\multicolumn{1}{l|}{Input}            & \cellcolor[HTML]{EFEFEF}\textit{11 channels x 72 datapoints} \\
\multicolumn{1}{l|}{\textit{Block 1}} & Conv1d (128 filters, kernel size 5, stride 2, Mish)          \\
\multicolumn{1}{l|}{\textit{}}        & \cellcolor[HTML]{EFEFEF}GroupNorm  (1 group, 128   channels) \\
\multicolumn{1}{l|}{\textit{Block 2}} & Conv1d (128 filters, kernel size 3, stride 2, Mish)          \\
\multicolumn{1}{l|}{\textit{}}        & \cellcolor[HTML]{EFEFEF}GroupNorm  (1 group, 128   channels) \\
\multicolumn{1}{l|}{\textit{Block 3}} & Conv1d (256 filters, kernel size 3, stride 2, Mish)          \\
\multicolumn{1}{l|}{\textit{}}        & \cellcolor[HTML]{EFEFEF}GroupNorm  (1 group, 256   channels) \\
\multicolumn{1}{l|}{\textit{}}        & Flatten                                                      \\
\multicolumn{1}{l|}{\textit{MLP}}     & \cellcolor[HTML]{EFEFEF}FC (9 * 256, 1, linear)                       \\
\multicolumn{1}{l|}{Output}           & \textit{1 value}                                            
\end{tabular}
\end{table}

%% file: Tables/GAN_table.tex
% Please add the following required packages to your document preamble:
% \usepackage[table,xcdraw]{xcolor}
% Beamer presentation requires \usepackage{colortbl} instead of \usepackage[table,xcdraw]{xcolor}
\begin{table}[ht!]
\centering
\caption{Description of our generator architecture using TCN-ResBlocks.}
\label{tab:TCN_Table}
\begin{tabular}{l}
\textbf{Generator Architecture}              \\ \hline
\textit{Input: 11 channels x 72 datapoints}  \\
\rowcolor[HTML]{EFEFEF} 
TCN-ResBlock(64, 3, 1, False)               \\
TCN-ResBlock(64, 3, 2, True)                \\
\rowcolor[HTML]{EFEFEF} 
TCN-ResBlock(64, 3, 4, True)                \\
TCN-ResBlock(64, 3, 8, True)                \\
\rowcolor[HTML]{EFEFEF} 
TCN-ResBlock(32, 3, 16, True)               \\
TCN-ResBlock(16, 3, 32, False)              \\
\rowcolor[HTML]{EFEFEF} 
1D Convolution (c = 11, k=1, d=1, stride=1, bias=True, linear)\\
\textit{Output: 11 channels x 72 datapoints}
\end{tabular}
\end{table}

%% file: Algorithms/algo1.tex
\begin{algorithm}[ht!]
\caption{Domain mapping network training algorithm.}
\label{alg:1}
\begin{algorithmic}[1]
\Require{Batch size $m$, generators $G_{ST}, G_{TS}$ and discriminators $Disc_S, Disc_T$, loss weight hyperparameters $\lambda_{cyc}, \lambda_{0}, \lambda_{R}$. In our experiments we set $\lambda_{cyc} = 30, \lambda_{0} = 0.5, \lambda_{R} = 0.1$.}
\While{Training not interrupted by early stopping}
\For{source domain batch $b_s$ and target domain batch $b_t$}

    \textbf{Generator updates}
    \State Map batches to corresponding other domain: $b_{st} = G_{ST}(b_s), b_{ts}=G_{TS}(b_t)$
    \State Map batches back to original domain: $b_{sts} = G_{TS}(b_st), b_{tst}=G_{ST}(b_ts)$
    \State $\mathcal{L}_{GAN_{ST}} \leftarrow \mathcal{L}_{GAN_{QP}}(b_t, b_{st}) $
    \State $\mathcal{L}_{GAN_{TS}} \leftarrow \mathcal{L}_{GAN_{QP}}(b_s, b_{ts}) $
    \State $\mathcal{L}_{GAN} \leftarrow \mathcal{L}_{GAN_{ST}} + \mathcal{L}_{GAN_{TS}} + \lambda_{cyc} \mathcal{L}_{cyc} + \lambda_{0} \mathcal{L}_0 + \lambda_R \mathcal{L}_R$
    \State Artificially corrupt batches $b_t$ and $b_s$; calculate and add $\lambda_{cyc} \mathcal{L}_{cyc}$ to $\mathcal{L}_{GAN}$
    \State Update weights $\mathbf{w}_{ST}, \mathbf{w}_{TS}$ of $G_{ST}, G_{TS}$ by descending:
    $\mathbf{w}_{TS}, \mathbf{w}_{ST} \leftarrow \text{Adam}(\nabla_{\mathbf{w}_{TS}, \mathbf{w}_{ST}} \mathcal{L}_{GAN})$

    \textbf{Discriminator updates}
    \State Sample new batches $b_s$ and $b_t$
    \State Map batches to corresponding other domain: $b_{st} = G_{ST}(b_s), b_{ts}=G_{TS}(b_t)$
    \State $\mathcal{L}_{Disc_S} \leftarrow \mathcal{L}_{Disc_{QP}}(b_s, b_{ts}) $
    \State $\mathcal{L}_{Disc_T} \leftarrow \mathcal{L}_{Disc_{QP}}(b_t, b_{st}) $
    \State $\mathcal{L}_{Disc} \leftarrow \mathcal{L}_{Disc_S} + \mathcal{L}_{Disc_T}$
    \State Update weights $\mathbf{w}_{S}, \mathbf{w}_{T}$ of $Disc_S, Disc_T$ by descending:
    $\mathbf{w}_{S}, \mathbf{w}_{T} \leftarrow \text{Adam}(\nabla_{\mathbf{w}_{S}, \mathbf{w}_{T}} \mathcal{L}_{Disc})$
\EndFor
\EndWhile
\end{algorithmic}
\end{algorithm}

%% file: Tables/f1tablefull.tex
% Please add the following required packages to your document preamble:
% \usepackage{booktabs}
% \usepackage{multirow}
% \usepackage{graphicx}
\begin{table}[ht!]
\centering
\caption{Detailed F1-scores (in \%) for all 28 domain pairs}
\label{tab:f1tablefull}
\resizebox{\textwidth}{!}{%
\begin{tabular}{@{}cl|cll|cll|cll|cll|cll|cll@{}}
\toprule
\multicolumn{1}{l}{} &
   &
  \multicolumn{3}{c|}{\textbf{2 months}} &
  \multicolumn{3}{c|}{\textbf{6   weeks}} &
  \multicolumn{3}{c|}{\textbf{1 month}} &
  \multicolumn{3}{c|}{\textbf{3 weeks}} &
  \multicolumn{3}{c|}{\textbf{2 weeks}} &
  \multicolumn{3}{c}{\textbf{1 week}} \\ \midrule
\multicolumn{1}{l}{\textbf{Target}} &
  \textbf{Source} &
  \textbf{Scarce} &
  \multicolumn{1}{c}{\textbf{FT}} &
  \multicolumn{1}{c|}{\textbf{Ours}} &
  \textbf{Scarce} &
  \multicolumn{1}{c}{\textbf{FT}} &
  \multicolumn{1}{c|}{\textbf{Ours}} &
  \textbf{Scarce} &
  \multicolumn{1}{c}{\textbf{FT}} &
  \multicolumn{1}{c|}{\textbf{Ours}} &
  \textbf{Scarce} &
  \multicolumn{1}{c}{\textbf{FT}} &
  \multicolumn{1}{c|}{\textbf{Ours}} &
  \textbf{Scarce} &
  \multicolumn{1}{c}{\textbf{FT}} &
  \multicolumn{1}{c|}{\textbf{Ours}} &
  \textbf{Scarce} &
  \multicolumn{1}{c}{\textbf{FT}} &
  \multicolumn{1}{c}{\textbf{Ours}} \\ \midrule
\multirow{4}{*}{WT01} &
  WT07 &
  \multirow{4}{*}{\textbf{89.0}} &
  \textbf{89.0} &
  87.9 &
  \multirow{4}{*}{84.6} &
  \textbf{88.5} &
  86.7 &
  \multirow{4}{*}{77.4} &
  77.5 &
  \textbf{86.1} &
  \multirow{4}{*}{\textbf{74.4}} &
  73.1 &
  57.3 &
  \multirow{4}{*}{77.1} &
  74.0 &
  \textbf{79.4} &
  \multirow{4}{*}{66.4} &
  74.7 &
  \textbf{78.8} \\
 &
  WT02 &
   &
  56.8 &
  64.4 &
   &
  77.4 &
  \textbf{86.3} &
   &
  72.2 &
  \textbf{89.5} &
   &
  70.2 &
  \textbf{79.6} &
   &
  72.9 &
  \textbf{83.2} &
   &
  72.3 &
  \textbf{76.7} \\
 &
  WT05 &
   &
  83.1 &
  82.2 &
   &
  80.9 &
  \textbf{88.3} &
   &
  73.7 &
  \textbf{84.0} &
   &
  75.1 &
  \textbf{87.9} &
   &
  79.0 &
  \textbf{83.6} &
   &
  75.2 &
  \textbf{75.5} \\
 &
  WT04 &
   &
  85.1 &
  \textbf{90.6} &
   &
  80.6 &
  \textbf{90.3} &
   &
  78.3 &
  \textbf{84.0} &
   &
  71.5 &
  \textbf{90.8} &
   &
  72.7 &
  \textbf{88.9} &
   &
  66.6 &
  \textbf{78.4} \\ \midrule
\multirow{4}{*}{WT02} &
  WT01 &
  \multirow{4}{*}{\textbf{77.8}} &
  \textbf{80.6} &
  50.0 &
  \multirow{4}{*}{62.6} &
  \textbf{83.7} &
  64.6 &
  \multirow{4}{*}{65.3} &
  \textbf{84.3} &
  56.3 &
  \multirow{4}{*}{60.3} &
  \textbf{73.9} &
  56.6 &
  \multirow{4}{*}{\textbf{73.6}} &
  32.7 &
  42.5 &
  \multirow{4}{*}{35.2} &
  1.9 &
  \textbf{41.8} \\
 &
  WT06 &
   &
  62.2 &
  71.7 &
   &
  \textbf{80.9} &
  69.0 &
   &
  \textbf{84.0} &
  80.0 &
   &
  74.9 &
  \textbf{77.1} &
   &
  40.7 &
  64.7 &
   &
  9.1 &
  \textbf{62.1} \\
 &
  WT05 &
   &
  \textbf{82.1} &
  63.5 &
   &
  \textbf{80.8} &
  64.3 &
   &
  \textbf{74.5} &
  65.2 &
   &
  \textbf{77.7} &
  65.5 &
   &
  \textbf{82.6} &
  61.2 &
   &
  \textbf{71.3} &
  61.4 \\
 &
  WT04 &
   &
  \textbf{84.7} &
  64.8 &
   &
  \textbf{83.9} &
  67.0 &
   &
  \textbf{72.7} &
  67.1 &
   &
  \textbf{78.6} &
  70.7 &
   &
  \textbf{80.7} &
  63.4 &
   &
  \textbf{66.0} &
  55.1 \\ \midrule
\multirow{6}{*}{WT03} &
  WT01 &
  \multirow{6}{*}{\textbf{84.7}} &
  36.1 &
  \textbf{86.5} &
  \multirow{6}{*}{73.8} &
  76.6 &
  \textbf{86.7} &
  \multirow{6}{*}{64.8} &
  56.9 &
  \textbf{84.8} &
  \multirow{6}{*}{38.4} &
  \textbf{76.8} &
  73.3 &
  \multirow{6}{*}{52.6} &
  67.1 &
  \textbf{73.5} &
  \multirow{6}{*}{34.5} &
  17.1 &
  \textbf{40.7} \\
 &
  WT06 &
   &
  43.2 &
  79.1 &
   &
  78.4 &
  \textbf{79.6} &
   &
  67.0 &
  \textbf{72.1} &
   &
  \textbf{75.2} &
  40.5 &
   &
  62.8 &
  \textbf{85.4} &
   &
  17.8 &
  \textbf{41.7} \\
 &
  WT07 &
   &
  53.2 &
  \textbf{86.8} &
   &
  71.7 &
  \textbf{85.6} &
   &
  \textbf{79.3} &
  48.7 &
   &
  \textbf{84.1} &
  46.0 &
   &
  76.3 &
  \textbf{89.3} &
   &
  \textbf{54.8} &
  50.2 \\
 &
  WT02 &
   &
  53.3 &
  78.4 &
   &
  78.9 &
  \textbf{83.4} &
   &
  \textbf{86.5} &
  48.7 &
   &
  73.7 &
  \textbf{76.2} &
   &
  66.5 &
  \textbf{78.6} &
   &
  41.0 &
  \textbf{52.5} \\
 &
  WT05 &
   &
  \textbf{88.8} &
  72.4 &
   &
  \textbf{89.9} &
  85.5 &
   &
  76.9 &
  \textbf{82.4} &
   &
  78.9 &
  \textbf{85.6} &
   &
  \textbf{88.7} &
  73.2 &
   &
  \textbf{90.8} &
  61.6 \\
 &
  WT04 &
   &
  81.7 &
  77.8 &
   &
  \textbf{89.8} &
  79.3 &
   &
  \textbf{74.8} &
  65.9 &
   &
  81.7 &
  \textbf{84.6} &
   &
  \textbf{85.9} &
  70.0 &
   &
  \textbf{65.1} &
  48.7 \\ \midrule
\multirow{5}{*}{WT04} &
  WT01 &
  \multirow{5}{*}{\textbf{92.6}} &
  79.8 &
  75.4 &
  \multirow{5}{*}{\textbf{94.0}} &
  70.9 &
  78.8 &
  \multirow{5}{*}{\textbf{88.3}} &
  78.8 &
  85.0 &
  \multirow{5}{*}{\textbf{89.8}} &
  67.4 &
  82.3 &
  \multirow{5}{*}{42.8} &
  42.3 &
  \textbf{54.5} &
  \multirow{5}{*}{34.4} &
  36.8 &
  \textbf{43.4} \\
 &
  WT06 &
   &
  86.3 &
  82.3 &
   &
  85.7 &
  72.3 &
   &
  81.0 &
  80.5 &
   &
  64.3 &
  86.7 &
   &
  45.3 &
  \textbf{49.7} &
   &
  42.3 &
  \textbf{47.0} \\
 &
  WT07 &
   &
  88.6 &
  84.3 &
   &
  92.3 &
  84.9 &
   &
  \textbf{89.9} &
  82.5 &
   &
  83.3 &
  78.9 &
   &
  60.4 &
  \textbf{66.3} &
   &
  57.3 &
  \textbf{61.3} \\
 &
  WT02 &
   &
  88.5 &
  92.3 &
   &
  81.1 &
  89.6 &
   &
  87.8 &
  \textbf{91.5} &
   &
  82.8 &
  85.4 &
   &
  63.3 &
  \textbf{69.9} &
   &
  52.4 &
  \textbf{61.5} \\
 &
  WT05 &
   &
  91.7 &
  \textbf{93.4} &
   &
  91.8 &
  93.5 &
   &
  92.8 &
  \textbf{94.7} &
   &
  92.1 &
  \textbf{93.5} &
   &
  \textbf{87.3} &
  61.1 &
   &
  \textbf{93.6} &
  63.4 \\ \midrule
\multirow{5}{*}{WT05} &
  WT01 &
  \multirow{5}{*}{\textbf{67.6}} &
  65.1 &
  61.9 &
  \multirow{5}{*}{55.3} &
  \textbf{64.9} &
  51.6 &
  \multirow{5}{*}{\textbf{63.9}} &
  53.8 &
  62.4 &
  \multirow{5}{*}{\textbf{56.5}} &
  46.4 &
  45.2 &
  \multirow{5}{*}{31.5} &
  26.0 &
  \textbf{33.2} &
  \multirow{5}{*}{24.3} &
  \textbf{31.9} &
  19.5 \\
 &
  WT06 &
   &
  65.8 &
  \textbf{78.3} &
   &
  63.5 &
  \textbf{76.6} &
   &
  54.8 &
  \textbf{70.7} &
   &
  50.0 &
  \textbf{80.8} &
   &
  34.1 &
  \textbf{78.7} &
   &
  13.1 &
  \textbf{47.4} \\
 &
  WT07 &
   &
  \textbf{87.4} &
  75.7 &
   &
  \textbf{83.5} &
  76.1 &
   &
  68.8 &
  \textbf{76.5} &
   &
  69.0 &
  \textbf{78.8} &
   &
  \textbf{64.5} &
  47.9 &
   &
  \textbf{60.8} &
  52.1 \\
 &
  WT02 &
   &
  \textbf{87.4} &
  81.8 &
   &
  \textbf{86.5} &
  80.7 &
   &
  83.2 &
  \textbf{86.1} &
   &
  76.5 &
  \textbf{88.9} &
   &
  \textbf{71.8} &
  62.9 &
   &
  49.4 &
  \textbf{53.1} \\
 &
  WT04 &
   &
  73.7 &
  \textbf{93.2} &
   &
  65.5 &
  \textbf{88.1} &
   &
  70.7 &
  \textbf{73.4} &
   &
  68.2 &
  \textbf{83.5} &
   &
  \textbf{82.5} &
  79.5 &
   &
  51.2 &
  \textbf{57.6} \\ \midrule
\multirow{4}{*}{WT06} &
  WT07 &
  \multirow{4}{*}{\textbf{94.5}} &
  67.3 &
  66.2 &
  \multirow{4}{*}{42.4} &
  59.1 &
  \textbf{60.8} &
  \multirow{4}{*}{24.7} &
  40.8 &
  \textbf{89.7} &
  \multirow{4}{*}{31.1} &
  36.3 &
  \textbf{49.6} &
  \multirow{4}{*}{30.3} &
  38.0 &
  \textbf{51.0} &
  \multirow{4}{*}{\textbf{7.0}} &
  6.4 &
  \textbf{10.9} \\
 &
  WT02 &
   &
  63.7 &
  76.7 &
   &
  60.5 &
  \textbf{73.6} &
   &
  42.5 &
  \textbf{74.5} &
   &
  39.3 &
  \textbf{47.0} &
   &
  32.5 &
  \textbf{40.1} &
   &
  6.1 &
  6.9 \\
 &
  WT05 &
   &
  74.9 &
  \textbf{94.9} &
   &
  42.9 &
  \textbf{57.8} &
   &
  20.7 &
  \textbf{54.0} &
   &
  18.3 &
  \textbf{75.7} &
   &
  19.5 &
  \textbf{70.0} &
   &
  6.2 &
  5.7 \\
 &
  WT04 &
   &
  79.4 &
  \textbf{94.9} &
   &
  59.7 &
  \textbf{76.5} &
   &
  28.1 &
  \textbf{72.3} &
   &
  15.9 &
  \textbf{75.8} &
   &
  21.5 &
  \textbf{78.9} &
   &
  6.0 &
  7.1 \\ \bottomrule
\end{tabular}%
}
\end{table}

%% file: Tables/datatable.tex
% Please add the following required packages to your document preamble:
% \usepackage{booktabs}
% \usepackage{graphicx}
\begin{table}[h!]
\centering
\caption{Data specifications of the 7 WTs used in our work. The number of days refers to the range between the day of the first sample and the day of the last sample, therefore including days where no (valid) measurements were taken.}
\label{tab:dataOverview}
\resizebox{\textwidth}{!}{%
\begin{tabular}{@{}l|lr|rr|rrr@{}}
\toprule
 &
   &
  \multicolumn{1}{l|}{} &
  \multicolumn{2}{c|}{\begin{tabular}[c]{@{}c@{}}Training \& Validation Set   \\ (filtered, without   scarcity)\end{tabular}} &
  \multicolumn{3}{c}{\begin{tabular}[c]{@{}c@{}}Test Set\\ (unfiltered, fixed)\end{tabular}} \\ \midrule
\textbf{WT\_ID} &
  \textbf{Location} &
  \multicolumn{1}{l|}{\textbf{Rated Power [KW]}} &
  \multicolumn{1}{l}{\textbf{\# days}} &
  \multicolumn{1}{l|}{\textbf{\# 12h-samples}} &
  \multicolumn{1}{l}{\textbf{\# days}} &
  \multicolumn{1}{l}{\textbf{\# 12h-samples}} &
  \multicolumn{1}{l}{\textbf{.. of which contain incidents}} \\
WT01 & Onshore & 800  & 899 & 50574 & 387 & 48736 & 9049 (18.6\%)  \\
WT02 & Onshore & 3000 & 877 & 58403 & 387 & 45593 & 19630 (43.1\%) \\
WT03 & Onshore & 2350 & 706 & 66407 & 323 & 43309 & 10102 (23.3\%) \\
WT04 & Onshore & 2050 & 831 & 70191 & 355 & 35573 & 3275 (9.2\%)   \\
WT05 & Onshore & 2300 & 827 & 70922 & 354 & 37308 & 788 (2.1\%)    \\
WT06 & Onshore & 800  & 798 & 70197 & 341 & 21187 & 121 (0.6\%)    \\
WT07 & Onshore & 3050 & 830 & 66117 & 355 & 43107 & 7775 (18.0\%)  \\ \bottomrule
\end{tabular}%
}
\end{table}